\documentclass[11pt,a4paper]{article}
\usepackage[hyperref]{An_Algorithm_for_Routing_Vectors_in_Sequences}
\usepackage{times}
\usepackage{latexsym}
\usepackage{graphicx}
\usepackage{caption}
\usepackage{amsmath}
\usepackage{amssymb}
\usepackage{amsfonts}
\usepackage{booktabs}
\usepackage[bottom]{footmisc}
\usepackage{url}
\usepackage[linesnumbered,lined,algoruled]{algorithm2e}
\usepackage{multicol}
\usepackage{qtree}
\graphicspath{{png/}}

\aclfinalcopy % Uncomment this line for the final submission
\title{	An Algorithm for Routing Vectors in Sequences }

\author{Franz A. Heinsen \\
	{\tt franz@glassroom.com} \\ }

\date{[Month Day], 2021}

\newcommand{\suptag}[1]{^{{\scriptscriptstyle\tt (#1)}}}
\newcommand{\iters}{\suptag{iters}}
\newcommand{\inp}{\suptag{inp}}
\newcommand{\out}{\suptag{out}}
\newcommand{\use}{\suptag{use}}
\newcommand{\ign}{\suptag{ign}}

\newcommand{\hid}{\suptag{hid}}
\newcommand{\emb}{\suptag{emb}}

\newcommand{\fnA}{\mathcal{A}}
\newcommand{\fnF}{\mathcal{F}}
\newcommand{\fnG}{\mathcal{G}}
\newcommand{\fnS}{\mathcal{S}}
\newcommand{\fnU}{\mathcal{U}}
\newcommand{\fnR}{\mathcal{R}}
\newcommand{\fnM}{\mathcal{M}}
\newcommand{\fnB}{\mathcal{B}}

\newcommand{\bigO}{\mathcal{O}}

\newcommand{\routing}{\mathsf{R}}

\newcommand{\commentstyle}[1]{\scriptsize\sffamily{#1}}
\newcommand{\textcomment}[1]{\text{\commentstyle{#1}}}

\SetCommentSty{commentstyle}

\begin{document}
\maketitle

\begin{abstract}
We propose a routing algorithm that takes a sequence of vectors and computes a new sequence with specified length and vector size. Each output vector maximizes ``bang per bit,'' the difference between a net benefit to use and net cost to ignore data, by better predicting the input vectors. We describe output vectors as geometric objects, as latent variables that assign credit, as query states in a model of associative memory, and as agents in a model of a Society of Mind. We implement the algorithm with optimizations that reduce parameter count, computation, and memory use by orders of magnitude, enabling us to route sequences of greater length than previously possible. We evaluate our implementation on natural language and visual classification tasks, obtaining competitive or state-of-the-art accuracy and end-to-end credit assignments that are interpretable.\footnote{Source code and instructions for replicating our results are online at \href{https://github.com/glassroom/heinsen_routing}{https://github.com/glassroom/heinsen\_routing}.}
\end{abstract}

\section{Introduction}\label{sec:introduction}

A longstanding goal in Artificial Intelligence is to formulate learning systems that assign credit, such that, when they succeed or fail in a task, we can determine and interpret which components of the system are responsible. A possible approach to the credit assignment problem is to route capsules at multiple levels of composition. A capsule is a group ({\em e.g.}, vector, matrix) of artificial neurons representing the properties of an entity in a context ({\em e.g.}, a token of text in a paragraph, an object depicted in an image). Routing consists of assigning data from input capsules, each representing a detected entity, to output capsules, each representing a detectable entity, by finding or computing agreement in some form ({\em e.g.}, identifying clusters) among candidate output capsules proposed by transforming the input capsules. Each output capsule is computed as a mixture of the candidates proposed for it on which the most input capsules agree, thereby assigning credit to those input capsules. If we compose multiple routings into a deep neural network, in every forward pass it assigns credit to input capsules representing the entities detected at each level of composition.

To date, deep neural networks applying various routing methods have shown promise in multiple domains, including vision and natural language, but only on small-scale tasks \cite{tsai2020Capsules} \cite{ribeiro2020capsule} \cite{NEURIPS2019_e46bc064} \cite{DBLP:journals/corr/abs-1902-05770} \cite{DBLP:journals/corr/abs-1911-00792} \cite{DBLP:journals/corr/abs-1904-09546} \cite{xinyi2018capsule} \cite{DBLP:journals/corr/abs-1805-10807} \cite{DBLP:journals/corr/abs-1805-10807} \cite{wang2018an} \cite{46653} \cite{DBLP:journals/corr/abs-1710-09829}. Application of previously proposed routing methods to large-scale tasks has been impractical due to computational complexity, which increases in both space and time as a function of the length of input and output sequences, the number of elements per capsule, and the number of pairwise interactions between input, proposed, and output capsules.

\begin{figure*}[t]
	\vskip 0.1in
	\begin{center}
		\centerline{\includegraphics{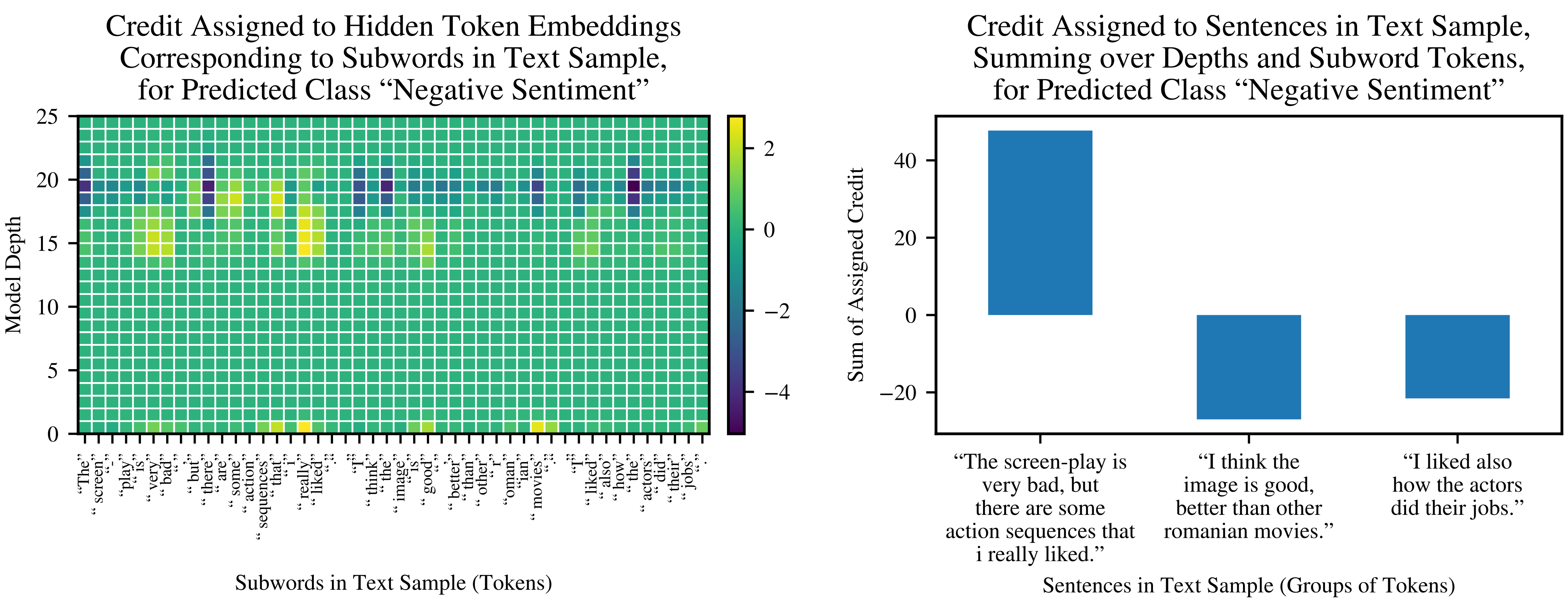}}
		\caption{Typical example of end-to-end credit assignment, in this case for classifying the sentiment of a movie review. See Figures \ref{fig:sample_credit_assignments_vision} and \ref{fig:sample_credit_assignments_natural_language} for additional examples, including a typical example in vision.}
		\label{fig:sample_credit_assignments_natural_language_for_introduction}
	\end{center}
	\vskip -0.2in
\end{figure*}

Here, we adapt the routing algorithm proposed by \citet{DBLP:journals/corr/abs-1911-00792} to operate on vectors as the capsules, generalize the algorithm by formulating it in terms of four neural networks (differentiable functions), and implement it with optimizations that reduce parameter count, memory use, and computation by orders of magnitude. The four neural networks are: $\fnA$ for obtaining an activation score per input vector, $\fnF$ for obtaining a different sequence of proposed output vectors given each input vector, $\fnG$ for predicting input vectors given a sequence of output vectors, and $\fnS$ for scoring actual versus predicted input vectors to quantify agreement. The algorithm is iterative. In each iteration, we update the state of all output vectors in parallel. We assign data from each input vector to the output vectors which best predict it, and compute each output vector's updated state by maximizing ``bang per bit,'' the difference between a net benefit to use and a net cost to ignore input vector data. The output sequence's final state is that which maximizes ``bang per bit'' by best predicting, or explaining, the given input sequence.

Motivated by consilience, we describe output vectors from four different viewpoints: First, we describe them as geometric objects whose states are updated by linearly combining dynamically computed coefficients and components in a different basis for each output vector. Second, we describe output vectors as latent variables whose updated states are computed via credit assignments that are additive, like the Shapley values obtainable via SHAP methods \cite{NIPS2017_7062}, and also composable on their own ({\em i.e.}, independently of data transformations), subject to certain conditions. Third, we describe output vectors as query states in a model of associative memory, which, if we disregard the net cost to ignore data and restrict  $\fnA$, $\fnF$, $\fnG$, and $\fnS$ in significant ways, reduces to a modern Hopfield network with the structure of a bipartite graph \cite{krotov2021large} \cite{ramsauer2021hopfield}, of which Transformer self-attention \cite{DBLP:journals/corr/VaswaniSPUJGKP17} is a notable special case. Fourth, we describe output vectors as agents competing to maximize utility by using or ignoring scarce resources through ``knowledge lines,'' or K-lines, in a ``block,'' which itself can interact with other blocks in a network modeling a Society of Mind \cite{10.5555/22939}.

Our sample implementation of the algorithm incorporates three significant optimizations: First, we define $\fnF$ as the composition of a scaled tensor product and a linear transformation with positionwise biases, instead of as a different linear transformation and bias per interaction (as in the original variant of the algorithm), reducing parameter count by orders of magnitude. Second, we evaluate $\fnF$ lazily in each iteration, instead of eagerly before the first iteration, to avoid storing all elements of $\fnF$'s output simultaneously in memory as intermediate values, reducing memory footprint by orders of magnitude while increasing computation only linearly in the number of iterations. Third, we decompose the computation of all updated output vector states into a sequence of efficient tensor contractions, reducing memory footprint and computation by orders of magnitude.

We measure our implementation's parameter count, memory footprint, and execution time, and find they are linear in each of the number of input vectors, the size of input vectors, the number of output vectors, and the size of output vectors, enabling fine-grained control over memory consumption and computational cost. We successfully route input sequences with 1 million vectors, each a capsule with 1024 elements, at full (32-bit floating point) precision, keeping track of gradients, consuming under 18GB of memory on widely available commodity hardware. To the best of our knowledge, no implementation of any previously proposed routing method has been able to route as many capsules on any kind of hardware.

Finally, we evaluate our implementation on classification benchmarks in natural language and vision. In all benchmarks, we obtain accuracy competitive with or better than the state of the art, along with additive credit assignments that are composable independently of data transformations. We compute end-to-end credit assignments and find they are interpretable (Figures \ref{fig:sample_credit_assignments_natural_language_for_introduction},  \ref{fig:sample_credit_assignments_vision}, \ref{fig:sample_credit_assignments_natural_language}).

\subsection{Notation}

In mathematical expressions of tensor transformations, we show all indices as subscript text, implicitly assume broadcasting for any missing indices, perform all operations elementwise, and explicitly show all summations. Superscript text in parenthesis denotes labels. See Table \ref{tab:notation_examples} for examples. We do not use the notation of Linear Algebra because it cannot handle more than two indices. We do not use Einstein's implicit summation notation because it would require the use of operators for raising and lowering indices, adding complexity that is unnecessary for our purposes.

\begin{table}[h]
	\small
	\begin{center}
		\begin{tabular}{@{}ll@{}}
			\toprule
			Example                                                              & Implementation in Python \\
			\midrule
			\addlinespace[0.6em]
			$y_{ijk} \longleftarrow x\suptag{1}_{ij} + x\suptag{2}_{jk}$         & {\tt y = x1[:,:,None] + x2} \\
			\addlinespace[0.6em]
			$y_{ijk} \longleftarrow x\suptag{1}_{ij} x\suptag{2}_{jk}$           & {\tt y = x1[:,:,None] * x2} \\
			\addlinespace[0.6em]
			$y_{ik} \longleftarrow \sum_j x\suptag{1}_{ij} x\suptag{2}_{jk}$     & {\tt y = x1 @ x2} \\
			\addlinespace[0.6em]
			$y_{ki} \longleftarrow e^{\sum_j x\suptag{1}_{ij} x\suptag{2}_{jk}}$ & {\tt y = (x1 @ x2).exp().T} \\
			\addlinespace[0.6em]
			$y_{k} \longleftarrow \sum_{ij} x\suptag{1}_{ij} x\suptag{2}_{jk}$   & {\tt y = (x1 @ x2).sum(dim=0)} \\
			\addlinespace[0.6em]
			\bottomrule
		\end{tabular}
	\end{center}
	\caption{\label{tab:notation_examples}Examples of the notation we use, with all-subscript tensor indices, elementwise operations, implicit broadcasting, and explicit summations. In all examples, $x\suptag{1}_{ij} \in \mathbb{R}^{d_1 \times d_2}$ and $x\suptag{2}_{jk} \in \mathbb{R}^{d_2 \times d_3}$.}
\end{table}

\section{Proposed Routing Algorithm}\label{sec:proposed_algorithm}

The proposed algorithm executes a modified expectation-maximization loop with three steps: an E-Step for computing expected routing probabilities, a D-Step for computing shares of data used and ignored, and an M-Step for computing output vectors that maximize ``bang per bit'' by more accurately predicting the given input vectors. The original variant of the algorithm \cite{DBLP:journals/corr/abs-1911-00792} routes matrices instead of vectors, in a loop with the same three steps, and computes output matrices as Gaussian mixtures that maximize the probability of generating the proposed ones, weighted by probabilities that maximize ``bang per bit.'' {\em For ease of exposition, we describe the new variant of the algorithm assuming the reader has no familiarity with the original one.}

\subsection{Overview}

\begin{algorithm*}[t]
	\small
	\DontPrintSemicolon
	\caption{$\fnA$, $\fnF$, $\fnG$, and $\fnS$ are implementation-specific. $f$ is the logistic function. $\beta\use_{ij}$ and $\beta\ign_{ij}$ are parameters if $n\inp$ is fixed, implementation-specific transformations of $x\inp_{id}$ otherwise.}
	\label{alg1:General_Formulation}
	\KwIn{$x\inp_{id} \in \mathbb{R}^{ n\inp \times d\inp }$.}
	\KwOut{$x\out_{jh} \in \mathbb{R}^{ n\out \times d\out }.$}
	\BlankLine
	$a\inp_i \longleftarrow \fnA(x\inp_{id}), \quad \fnA : \mathbb{R}^{ n\inp \times d\inp } \to \mathbb{R}^{ n\inp }$ \tcp*{obtain input vector activation scores} \label{alg1:a_inp_i}
	$V_{ijh} \longleftarrow
	\fnF \left( x\inp_{id} \right), \quad \fnF : \mathbb{R}^{ n\inp \times d\inp } \to \mathbb{R}^{ n\inp \times n\out \times d\out }$ \tcp*{obtain votes (proposed output vectors)} \label{alg1:V_ijh}
	\For{$n\iters$ iterations}{
		\Begin(E-Step){
			\eIf{on first iteration}{
				$R_{ij} \longleftarrow \frac{1}{n\out}$ \tcp*{assign flat prior in first iteration} \label{alg1:initial_R_ij}
			}{
				$\hat{x}\inp_{jd} \longleftarrow \fnG(x\out_{jh}),
				\quad \fnG : \mathbb{R}^{ n\out \times d\out } \to \mathbb{R}^{ n\out \times d\inp }$ \tcp*{predict input vectors} \label{alg1:x_inp_hat_jd}
				$S_{ij} \longleftarrow \fnS(x\inp_{id},\hat{x}\inp_{jd}),
				\quad \fnS : \mathbb{R}^{ n\inp \times d\inp } \times \mathbb{R}^{ n\out \times d\inp } \to \mathbb{R}^{ n\inp \times n\out }$
				\tcp*{score the predictions} \label{alg1:S_ij}
				$R_{ij} \longleftarrow \frac{ e^{S_{ij}} }{ \sum_j e^{S_{ij}} }$ \tcp*{normalize to distributions} \label{alg1:R_ij}
			}
		}
		\Begin(D-Step){
			$D\use_{ij} \longleftarrow f(a\inp_i) R_{ij}$ \tcp*{compute shares of data used} \label{alg1:D_use_ij}
			$D\ign_{ij} \longleftarrow f(a\inp_i) - D\use_{ij}$ \tcp*{compute shares of data ignored} \label{alg1:D_ign_ij}
		}
		\Begin(M-Step){
			$x\out_{jh} \longleftarrow \sum_i \beta\use_{ij} D\use_{ij} V_{ijh} - \sum_i \beta\ign_{ij} D\ign_{ij} V_{ijh}$ \tcp*{maximize ``bang per bit''} \label{alg1:x_out}
		}
	}
\end{algorithm*}

We show the proposed algorithm as Algorithm \ref{alg1:General_Formulation}. Per sample, we accept a sequence of input vectors $x\inp_{id}$ and return a sequence of output vectors $x\out_{jh}$. The tensor indices, which we use consistently throughout the rest of this document, are:

\begin{equation*}
\begin{aligned}
i & = (1, 2, \dots, n\inp), \\
j & = (1, 2, \dots, n\out), \\
d & = (1, 2, \dots, d\inp), \\
h & = (1, 2, \dots, d\out), \\
\end{aligned}
\end{equation*}

where $n\inp$ and $n\out$ are the number of input and output vectors, respectively, and $d\inp$ and $d\out$ are the size, or number of features, of input and output vectors, respectively.

In the following subsections, we walk through all steps of Algorithm \ref{alg1:General_Formulation} in order of execution.

\subsection{Input Vector Activations}\label{ssec:a_inp_i}

We apply a neural network $\fnA$ to the input vectors to obtain their activation scores $a\inp_i$ (Algorithm \ref{alg1:General_Formulation}, line \ref{alg1:a_inp_i}), and subsequently apply a logistic function $f$ to each activation score to obtain a probability per input vector $f(a\inp_i)$ (Algorithm \ref{alg1:General_Formulation}, lines \ref{alg1:D_use_ij}-\ref{alg1:D_ign_ij}).\footnote{
	We represent the logistic function with $f$ instead of $\sigma$, as is conventional, because the latter denotes standard deviation elsewhere in this document and in the original algorithm.
} We call each such probability an ``input vector activation'' and use it to gate the input vector's proposed output vectors.

A notable special case of $\fnA$, which we will revisit, is defining it as a constant function, $\fnA(\cdot) := \infty$, making all input vector activations $f(\infty) = 1$, in which case we always activate all ({\em i.e.}, never gate any) proposed output vectors.

\subsection{Proposed Output Vectors, or Votes}\label{ssec:V_ijh}

We apply a neural network $\fnF$ to the input vectors to obtain a tensor of proposed output vectors $V_{ijh}$ (Algorithm \ref{alg1:General_Formulation}, line \ref{alg1:V_ijh}). The tensor $V_{ijh}$ has, for each input vector $i$, a proposed vector for each possible output vector $j$ with features $h$. We call each proposed output vector a ``vote'' to distinguish it from actual output vectors, which we compute at the end of each iteration in the routing loop. $\fnF$ {\em should break symmetry}, {\em i.e.}, obtain from each input vector a different vote for each output vector; otherwise, all votes from each input vector $i$ would be identical and routing would be pointless.

A notable special case of $\fnF$, which we will revisit, is defining it as a constant function that returns a parameter, $\fnF(\cdot) := W\suptag{mem}_{ijh}$, making all votes ``learnable memories'' that are independent of the given input vectors, retrieved instead of computed from them at inference.

Implementing $\fnF$ presents two difficulties to routing long sequences. First, storing the votes $V_{ijh}$ requires $\bigO(n\inp n\out d\out)$ space, which becomes impractical as we increase the length of input and output sequences. Second, naive approaches to breaking symmetry require parameter counts that also become impractical as we increase the length of input and output sequences. For example, applying $n\inp n\out$ different linear transformations (as in the original variant of the algorithm) would require $n\inp n\out d\inp d\out$ parameters, and $n\out$ linear transformations would require $n\out d\inp d\out$ parameters. In Section \ref{sec:implementation}, we present a sample implementation with optimizations that overcome both difficulties, by lazily computing, weighting, and contracting $V_{ijh}$'s elements without storing all of them simultaneously as intermediate values, in an efficient manner. For now, we set aside concerns about routing longer sequences and focus on the next step of the algorithm: the routing loop.

\subsection{Routing Loop}\label{ssec:routing_iterations}

\subsubsection{E-Step}\label{sssec:E_Step}

The E-Step computes a tensor $R_{ij}$ of expected routing probabilities, for assigning data from each input vector $i$'s votes to compute each output vector $j$. In the first iteration, we assign equal routing probability, $\frac{1}{n\out}$, {\em i.e.}, a flat prior, over the votes from each input vector (Algorithm \ref{alg1:General_Formulation}, line \ref{alg1:initial_R_ij}).

In subsequent iterations, we assign greater routing probability to the output vector states which best predict the input vectors, as follows: First, we predict input vectors by applying a neural network $\fnG$ to the the previous iteration's output vector states (line \ref{alg1:x_inp_hat_jd}). We obtain $n\out$ predicted input vectors. Second, we compute prediction scores $S_{ij}$ by applying a neural network $\fnS$ to actual and predicted input vectors (line \ref{alg1:S_ij}). $\fnS$ may compute a symmetric kernel (dot-product, Euclidean distance, radial basis function, etc.) or a non-symmetric kernel ({\em i.e.}, one that computes scores differently for different pairs of actual and predicted input vectors, breaking symmetry over the input vectors too).\footnote{
	Optionally, $\fnG$ may specify a generative model that samples the predicted input vectors given current output vector states, in which case $\fnS$ should compute or approximate the conditional log-probability densities of actual input vectors, given the predicted input vectors, as the scores $S_{ij}$.
} Finally, we apply a Softmax function to $S_{ij}$, normalizing over index $j$, to obtain updated routing probabilities $R_{ij}$ which add up to 1 per input vector; {\em i.e.}, for each input vector $i$ we obtain a distribution over the input vector's proposed output vector states $j$ (line \ref{alg1:R_ij}).

\subsubsection{D-Step}\label{ssec:D_Step}

The D-Step computes the shares of data used $D\use_{ij}$ and ignored $D\ign_{ij}$ from each input vector $i$'s vote for computing the state of each output vector $j$. We use these shares to put output vectors in competition with each other as they try to use ``more valuable bits'' and ignore ``less valuable bits'' of data from each input vector's votes, such that each output vector can use more data from an input vector's votes only if all other output vectors collectively ignore it, and vice versa.

We obtain $D\use_{ij}$ by multiplying input vector activations $f(a\inp_i)$ by routing probabilities $R_{ij}$ (line \ref{alg1:D_use_ij}). Each element of $f(a\inp_i)$ is in $[0,1]$ and the elements of $R_{ij}$ along index $j$ add up to 1, so the elements of $D\use_{ij}$ have values that range from 0 (``ignore all data from input vector $i$'s vote for output vector $j$'') to 1 (``use all data from input vector $i$'s vote for output vector $j$''), but never exceed each input vector activation (``how much data from input vector $i$'s votes can all output vectors collectively use?''). We then compute $D\ign_{ij}$ by subtracting the shares used from the input vector activations (line \ref{alg1:D_ign_ij}), such that for every input vector $i$ and every output vector $j$,

\begin{equation}\label{eq:sum_of_shares_equals_f_a_inp}
\begin{aligned}
D\use_{ij} + D\ign_{ij} & = f(a\inp_i) \\
\sum_j D\use_{ij} & = f(a\inp_i), \\
\end{aligned}
\end{equation}

where

\begin{equation}\label{eq:each_share_le_f_a_inp}
\begin{aligned}
0 \le D\use_{ij} & \le f(a\inp_i) \le 1 \\
0 \le D\ign_{ij} & \le f(a\inp_i) \le 1, \\
\end{aligned}
\end{equation}

treating activated (non-gated) data as a scarce resource that cannot be wasted: Every bit must be ``fully used'' by one or more output vectors and ``fully ignored'' by all other output vectors.

\subsubsection{M-Step}

The M-Step computes updated output vector states $x\out_{jh}$ at the end of each iteration as the difference between each output vector's net benefit to use and net cost to ignore data from input vector votes, maximizing ``bang per bit'' (line \ref{alg1:x_out}). The word ``net'' denotes that values may be positive or negative---{\em i.e.}, it is possible for the net benefit to be negative and for the net cost to be positive.

We compute each output vector's net benefit to use data as a linear combination of the votes, where the coefficients are the shares of data used, scaled by a parameter $\beta\use_{ij}$ quantifying each output vector's net benefit per unit of data to use each vote---hence the term ``bang per bit.'' We compute the net cost to ignore data also as a linear combination of the votes, where the coefficients are the shares of data ignored, scaled by a parameter $\beta\ign_{ij}$ quantifying each output vector's net cost per unit of data to ignore each vote.

For example, the first output vector's state is

\begin{equation}\label{eq:x_out_1h}
\begin{aligned}
x\out_{1h} \longleftarrow \phantom{-}
& \sum_i \beta\use_{i1} D\use_{i1} V_{i1h}
& \textcomment{// net benefit} \\
-
& \sum_i \beta\ign_{i1} D\ign_{i1} V_{i1h},
& \textcomment{// net cost}
\end{aligned}
\end{equation}

where the tensor slice $V_{i1h}$ has the votes from input vectors $i$ for output vector 1 with elements $h$, $D\use_{i1}$ and $D\ign_{i1}$ are the shares of data from each input vector $i$ used and ignored by output vector 1, and $\beta\use_{i1}$ and $\beta\ign_{i1}$ are the net benefit and net cost per unit of data from input vector $i$ for output vector 1. We maximize the first output vector's net benefit from those votes it uses, less its net cost from those votes it ignores, in competition with all other output vectors, for which we do the same.

If no output vector can improve its net benefit less net cost, given the state of all other output vectors, the routing loop has reached a local optimum in a ``bang per bit'' landscape, specific to the implementation of neural networks $\fnA$, $\fnF$, $\fnG$, and $\fnS$, given the current sequence of input vectors.

\subsection{Training}\label{ssec:training}

We optimize all parameters for a training objective specified elsewhere as a dependency of the output vector states, which in turn are a function of (a) $\beta\use_{ij}$ and $\beta\ign_{ij}$, (b) $D\use_{ij}$ and $D\ign_{ij}$, and (c) $V_{ijh}$ in each iteration. Provided the implementation of $\fnA$, $\fnF$, $\fnG$, and $\fnS$ exhibits Lyapunov stability in the routing loop, we can directly optimize (a), which are learnable parameters, and (c), the votes, which are proposed by a differentiable function ($\fnF$), but not (b), which we can optimize only indirectly, via the interaction of input vector activations $f(a\inp_{i})$ and actual-versus-predicted input vector scores $S_{ij}$, which together determine $D\use_{ij}$ and $D\ign_{ij}$, subject to  \eqref{eq:sum_of_shares_equals_f_a_inp} and \eqref{eq:each_share_le_f_a_inp}, {\em inducing the algorithm to learn to activate and predict input vectors} as we optimize for the training objective.

If the training objective induces each output vector's elements to represent the properties of an object, concept, relationship, or other entity for which we, human beings, already have a label, each output vector is {\em a symbol for a known entity}. Otherwise, the algorithm learns to compute output vector states representing objects, concepts, relationships, or other entities for which we, human beings, may or may not have labels ({\em e.g.}, we may be unaware of their existence), making each output vector {\em a symbol for a discoverable entity}.

\section{Understanding Output Vectors}\label{sec:understanding_output_vectors}

\subsection{As Geometric Objects}

If we factorize out $V_{ijh}$ from the expression in line \ref{alg1:x_out} of Algorithm \ref{alg1:General_Formulation}, we see that each iteration computes the updated state of each output vector as the linear combination of a vector basis in $V_{ijh}$ with corresponding ``bang per bit'' coefficients $\phi_{ij}$:

\begin{equation}\label{eq:phi_ij}
\begin{aligned}
x\out_{jh} \longleftarrow & \sum_i (\underbrace{ \beta\use_{ij} D\use_{ij} - \beta\ign_{ij} D\ign_{ij}}_{
	\textcomment{Define as $\phi_{ij}$}
}) V_{ijh} \\
= & \sum_i
\underbrace{ \phi_{ij} }_{\begin{subarray}{c}
	\textcomment{Coeffi-} \\
	\textcomment{cients} \\
	\end{subarray}}
\underbrace{ V_{ijh} }_{\begin{subarray}{c}
	\textcomment{Vector} \\
	\textcomment{bases} \\
	\end{subarray}}.
\end{aligned}
\end{equation}

Neural network $\fnF$ transforms input vectors into {\em a different basis for each output vector}. The tensor $V_{ijh}$ consists of $n\inp$ votes specifying a basis for each of $n\out$ output vectors of size $d\out$.\footnote{For intuition's sake, we can think of each vote as a basis vector, even though the vote is properly a basis vector only if it is linearly independent of all other votes in the same basis.} In the special case where $\fnF$ is a constant function that returns a parameter $W\suptag{mem}_{ijh}$, every basis is a learned memory, retrieved given the input vectors instead of computed from them at inference. Each basis may represent a different feature space.

\begin{figure}[t]
	\vskip 0.1in
	\begin{center}
		\centerline{\includegraphics{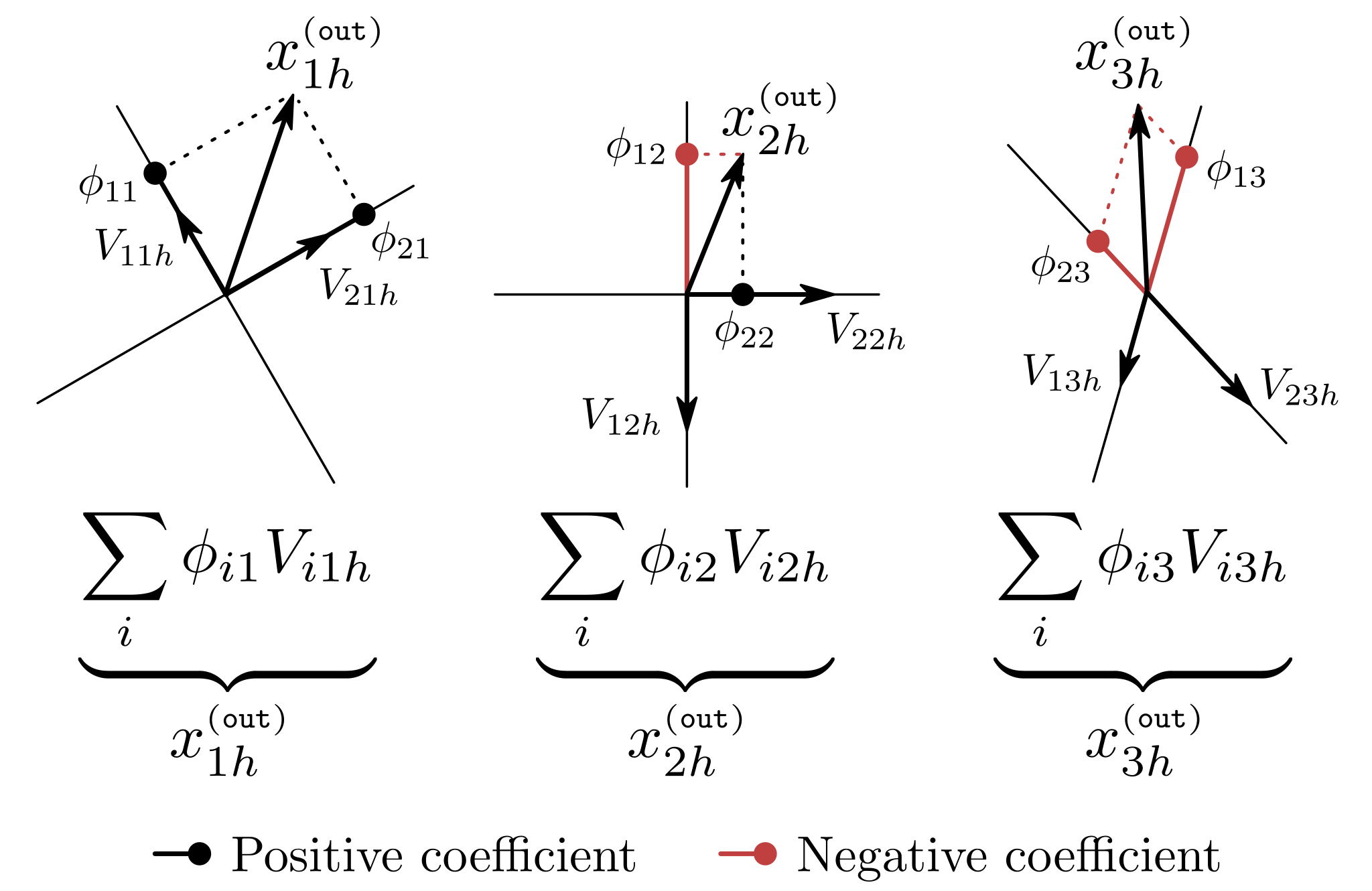}}
		\caption{Each output vector $j$'s state is the linear combination of its corresponding basis $ih$ in $V_{ijh}$ with its dynamically updated coefficients $i$ in $\phi_{ij}$. In this illustration, $n\inp=2$ and $n\out=3$.}
		\label{fig:as_geometric_objects}
	\end{center}
	\vskip -0.2in
\end{figure}

For example, the computation of the first output vector's state $x\out_{1h}$ in \eqref{eq:x_out_1h} is factorized as

\begin{equation}\label{eq:x_out_1h_with_phi}
x\out_{1h} \longleftarrow \sum_i \phi_{i1} V_{i1h}
\end{equation}

where the tensor slice $V_{i1h}$ is the basis specified by votes $i$ for output vector 1 with elements $h$, and tensor slice $\phi_{i1}$ has the coefficients $i$ for the votes that specify output vector 1's basis. Figure \ref{fig:as_geometric_objects} illustrates an example with three bases ($V_{i1h}, V_{i2h}, V_{i3h}$) and three slices with coefficients ($\phi_{i1}, \phi_{i2}, \phi_{i3}$) obtained from two input vectors for computing the state of three output vectors.

When we maximize ``bang per bit,'' we find the coordinates in each basis that best predict the input vectors, subject to constraints \eqref{eq:sum_of_shares_equals_f_a_inp} and \eqref{eq:each_share_le_f_a_inp}, in service of a training objective, specified elsewhere as a dependency of the final output vector states.

\subsection{As Latent Variables that Assign Credit}

\begin{figure}[t]
	\vskip 0.1in
	\begin{center}
		\centerline{\includegraphics{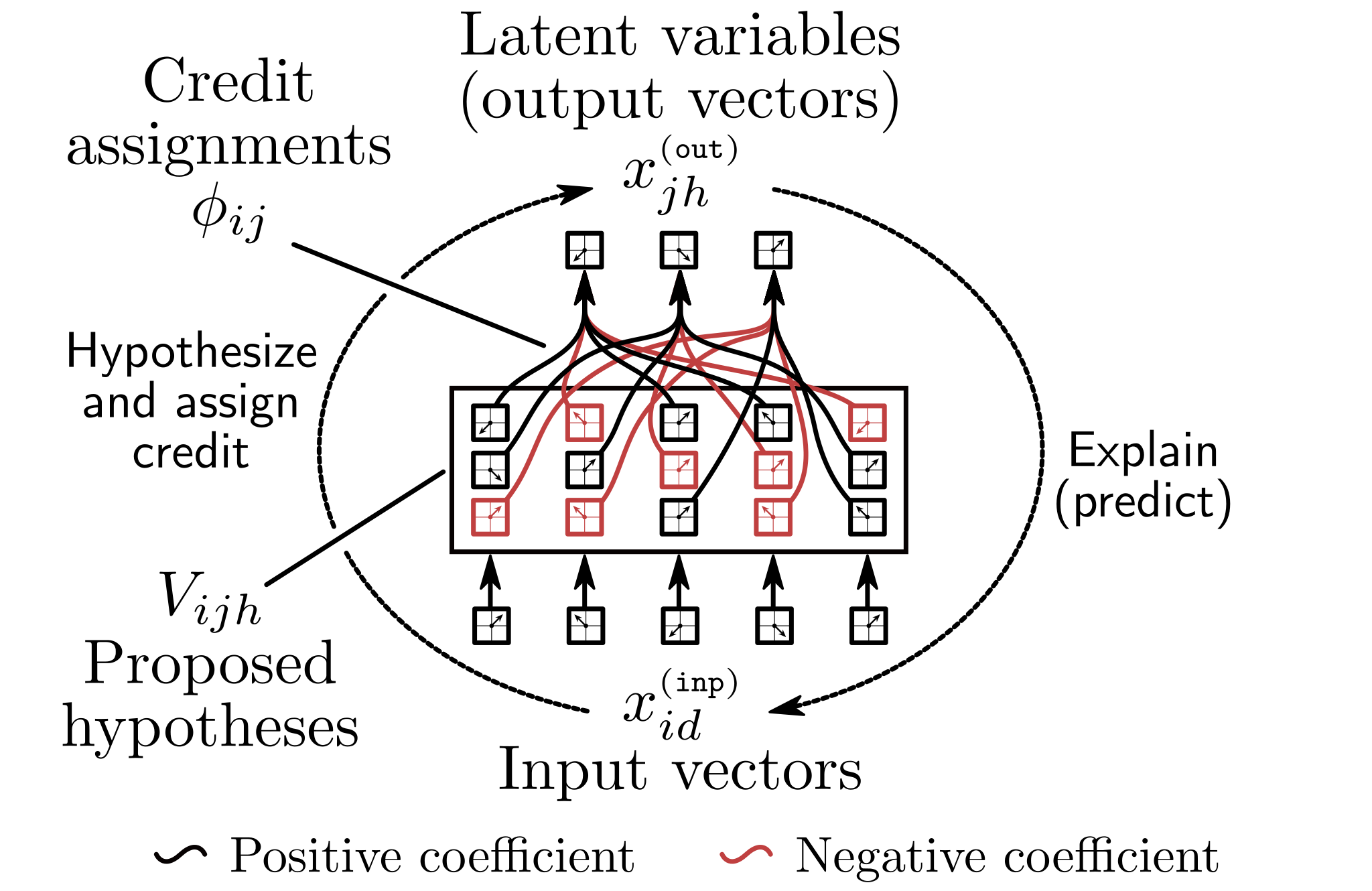}}
		\caption{We find the credit assignments in each space of proposed hypotheses that compute the output vector states which best explain the input vectors. In this diagram, $n\inp = 5$ and $n\out = 3$.}
		\label{fig:latent_variables}
	\end{center}
	\vskip -0.2in
\end{figure}

We can describe each output vector as a latent or explanatory variable with $d\out$ elements. From this viewpoint, each basis in $V_{ijh}$ is a space of ``proposed hypotheses'' for one output vector. Different coordinates in each basis represent different hypotheses for explaining, or predicting, the given sequence of input vectors (Figure \ref{fig:latent_variables}). In the special case where $\fnF$ is a constant function that returns a parameter $W\suptag{mem}_{ijh}$, every space of proposed hypotheses is a learned memory, retrieved instead of computed from the input vectors at inference.

The ``bang per bit'' coefficients $\phi_{ij}$ \eqref{eq:phi_ij}, or coordinates in each space of proposed hypotheses, specify how much each input vector $i$'s proposed hypothesis adds to, or subtracts from, each output vector $j$'s updated state. That is, the coefficients {\em assign credit via addition and subtraction} of each input vector $i$'s proposed hypothesis to compute each output vector $j$'s updated state. Compared to SHAP methods \cite{NIPS2017_7062}, which estimate additive credit assignments by sampling model outputs on a sufficiently large number of perturbations applied to a given input sample, our algorithm gives us additive credit assignments ``for free'' via an iterative forward pass, without having to figure out how best to perturb input data.

From this viewpoint, maximizing ``bang per bit'' means finding the credit assignments $\phi_{ij}$ in all spaces of proposed hypotheses $V_{ijh}$, for computing the output vector states $x\out_{jh}$ which best explain the sequence of input vectors $x\inp_{id}$, in service of a training objective specified elsewhere as a dependency of the final output vector states. If no output vector can improve its predictions, given the state of all other output vectors, the algorithm has reached a local credit-assignment optimum in a landscape of proposed hypotheses specific to the implementation of $\fnA$, $\fnF$, $\fnG$, and $\fnS$.

\begin{figure}[t]
	\vskip 0.1in
	\begin{center}
		\centerline{\includegraphics{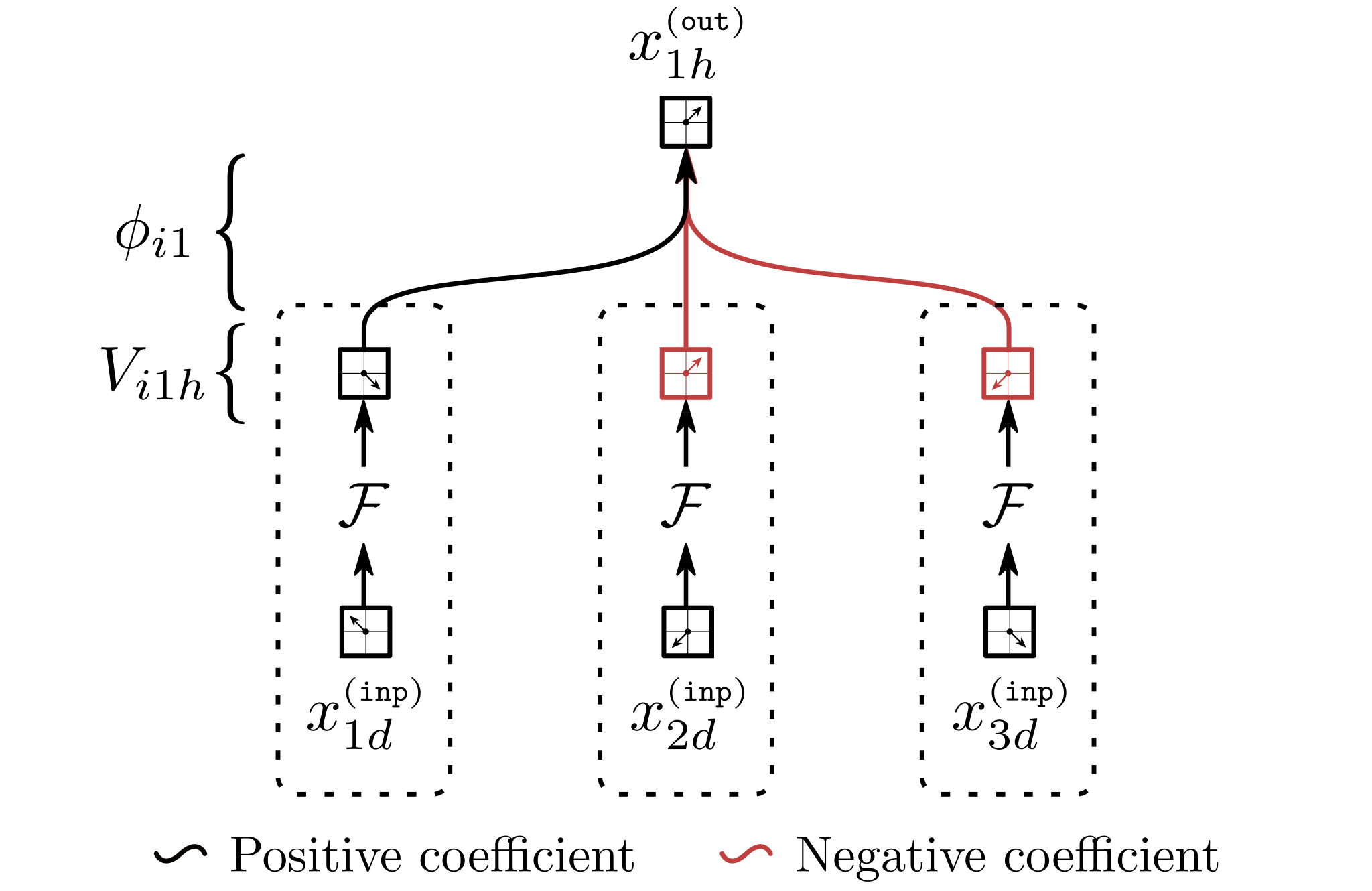}}
		\caption{If $\fnF$ keeps data from each input vector separate, then data from different input vectors is mixed only by $\phi_{ij}$. Here, we show three independently obtained votes for the first output vector.}
		\label{fig:composable_credit_assignments}
	\end{center}
	\vskip -0.2in
\end{figure}

If we implement $\fnF$ to obtain each input vector's votes independently of other input vectors' votes, then data from different input vectors is mixed only by $\phi_{ij}$ (Figure \ref{fig:composable_credit_assignments}), making the credit assignments composable on their own, independently of data transformations: In a network of routings, data from different vectors is mixed only by the final credit assignments computed by each routing. In appendix \ref{app:composability_of_credit_assignments}, we show methods for computing end-to-end credit assignments over common compositions of routings, including residual layers.

\subsection{As Associative Memory Query States}\label{ssec:as_associative_memory_states}

We can describe the proposed algorithm as applying an update rule $\fnU$ to output vectors in each iteration, given a sequence of input vectors:

\begin{equation}
\underbrace{x\out_{jh}}_{\textcomment{Updated}} \longleftarrow \fnU ( x\out_{jh} | x\inp_{id} ),
\end{equation}

where $\fnU$ composes all transformations we apply to output vectors in the E-Step, D-Step, and M-Step after the first iteration (lines \ref{alg1:x_inp_hat_jd}-\ref{alg1:x_out}). Grouping all such transformations into three newly defined neural networks, which we call $\fnR$, $\fnM$, and $\fnB$,\footnote{See appendix \ref{app:derivation_of_update_rule} for the derivation of $\fnU$ in terms of these three newly defined neural networks: $\fnR$, $\fnM$, and $\fnB$.} we see that $\fnU$ is the update rule for a model of associative memory with the structure of a bipartite graph in which output vectors are query states and input vectors are keys to content-addressable ``memory values'' and ``memory biases:''

\begin{equation}\label{eq:U_definition}
\begin{aligned}
& \fnU (x\out_{jh} | x\inp_{id} ) :=  \\
& \quad \sum_i \Big( \fnR ( \underbrace{ x\out_{jh} }_{\textcomment{Queries}} | \underbrace{ x\inp_{id\phantom{j}} }_{\textcomment{Keys}} )
\underbrace{ \fnM(x\inp_{id\phantom{j}}) }_{\textcomment{Values}}
- \underbrace{ \fnB(x\inp_{id\phantom{j}}) }_{\textcomment{Biases}} \Big), \\
\end{aligned}
\end{equation}

where $\fnR$ applies $\fnG$ and $\fnS$ to obtain updated routing probabilities $R_{ij}$ (E-Step, lines \ref{alg1:x_inp_hat_jd}-\ref{alg1:R_ij}),

\begin{equation}
\fnR( x\out_{jh} \, | \, x\inp_{id} ) := 
\frac{ e^{\fnS(x\inp_{id},\, \fnG( x\out_{jh} ))} }{ \sum_j e^{\fnS(x\inp_{id},\, \fnG( x\out_{jh} ))} },
\end{equation}

and $\fnM$ and $\fnB$ compose and weight the application of $\fnA$ and $\fnF$ in the D-Step and M-Step to obtain memory values and biases for each key,

\begin{equation}
\begin{aligned}
\fnM(x\inp_{id}) & := (\beta\use_{ij} \! + \! \beta\ign_{ij}) f(\fnA(x\inp_{id})) \fnF(x\inp_{id}) \\
\fnB(x\inp_{id}) & := \beta\ign_{ij} f(\fnA(x\inp_{id})) \fnF(x\inp_{id}), \\
\end{aligned}
\end{equation}

{\em i.e.}, $\fnM$ and $\fnB$ compute different scalings of the input vector votes, $\fnF(x\inp_{id})$, gated by corresponding input vector activations, $f(\fnA(x\inp_{id}))$.

In the special case where $\fnF$ is a constant function that returns a parameter $W\suptag{mem}_{ijh}$, the votes are learned memories, retrieved instead of computed from the keys, and $\fnM$ and $\fnB$ compute different gated scalings of such retrieved memories. If $\fnA$ is a constant function that returns $\infty$, all memories are always fully activated ({\em i.e.}, never gated).

The initial query states assign equal prior routing probability, $\frac{1}{n\out}$ (E-Step, line \ref{alg1:initial_R_ij}), to their corresponding memory values given each key:

\begin{equation}\label{eq:U_initial_state}
\underbrace{x\out_{jh}}_{\textcomment{Initial}} \longleftarrow \sum_i \Big( \underbrace{\mathsmaller{\frac{1}{n\out}}}_{\textcomment{Prior}}
\underbrace{ \fnM(x\inp_{id\phantom{j}}) }_{\textcomment{Values}}
- \underbrace{ \fnB(x\inp_{id\phantom{j}}) }_{\textcomment{Biases}} \Big).
\end{equation}

When we maximize ``bang per bit,'' we iteratively update query states as the mixtures of memory values, less memory biases, which best predict the given keys, in service of a training objective that we specify elsewhere as a dependency of the final query states. If no query can improve its predictions, given the state of all other queries, we have reached a local maximum in a ``bang per bit'' landscape (or equivalently, a local minimum in an energy landscape) specific to the implementation of neural networks $\fnA$, $\fnF$, $\fnG$, and $\fnS$.

Provided the implementation exhibits Lyapunov stability, we can view the algorithm as an ``infinitely deep'' recurrent neural network that repeatedly applies the same layer $\fnU$ to the queries until they converge to a stable state $\breve{x}\out_{jh}$:

\begin{equation}\label{eq:U_fixed_point}
\begin{aligned}
\breve{x}\out_{jh}
& = \lim_{t \to \infty} \fnU^t ( x\out_{jh} | x\inp_{id} ) \\
& = \fnU ( \breve{x}\out_{jh} | x\inp_{id} ), \\
\end{aligned}
\end{equation}

where $\fnU^t ( x\out_{jh} | x\inp_{id} )$ denotes $t$ applications of $\fnU$ to the queries, given the keys. Alternatively, we can think of the algorithm as a ``single layer'' implicitly defined by its output, the stable state $\breve{x}\out_{jh}$ that solves \eqref{eq:U_fixed_point}, making the algorithm a ``deep equilibrium model'' \cite{DBLP:journals/corr/abs-1909-01377}.\footnote{
	We consider only query states that evolve over a discrete number of iterations. Were we to extend our algorithm to the continuous setting, query states would evolve instead over time $t$ by a system of ordinary differential equations:
	\begin{equation*}
	\frac{\partial}{\partial t} x\out_{jh} (t) = \fnU' \left( x\out_{jh} (t) \, \big| \, x\inp_{id} \right),
	\end{equation*}
	with initial condition at $t=t_0$ given by an uniform prior distribution over each query's corresponding memory values given each key \eqref{eq:U_initial_state}. Alas, absent an analytical solution (or plausible implementation as a continuous physical process), we would have to approximate integration with numerical methods, requiring a discrete number of iterations anyway.
}

We believe our algorithm is the first model of associative memory to take into account a net cost to ignore data. If we simplify the algorithm, it reduces to a modern Hopfield network with bipartite structure \cite{krotov2021large} \cite{ramsauer2021hopfield}, of which Transformer self-attention \cite{DBLP:journals/corr/VaswaniSPUJGKP17} is a notable special case. The necessary simplifications are: (a) We would have to disregard the net cost to ignore data, {\em e.g.}, by restricting $\beta\ign_{ij}$ to constant 0, eliminating the memory biases obtained by $\fnB$ from expressions \eqref{eq:U_definition} and \eqref{eq:U_initial_state}. (b) We would have to restrict $\beta\use_{ij}$ to constant 1, so as to avoid scaling votes differently for each pair of input and output vectors. (c) We would have to restrict $\fnA$ to a constant function that returns $\infty$, always fully activating all ({\em i.e.}, never gating any) memory values obtained by $\fnM$. (d) We would have to restrict $\fnF$ (and thus $\fnM$) to propose only one sequence of proposed output vectors, {\em i.e.}, not to break symmetry over them, making routing unnecessary, and apply instead attention over that single sequence of proposed output vectors. (e) We would have to restrict $\fnM$ (which composes $\fnA$ and $\fnF$) and $\fnR$ (which composes $\fnG$ and $\fnS$) to those transformations guaranteed to converge to local optima proposed by \citet{krotov2021large} and \citet{ramsauer2021hopfield}.

\subsection{As Agents in a Society of Mind}

Output vectors are multidimensional agents competing with each other to use or ignore data representing input vectors. Each input vector is a scarce resource that cannot be wasted, as we account for all data, ensuring each agent can use or ignore it only at the expense of other agents, as described in \ref{ssec:routing_iterations}. Agents improve their use and ignore shares by more accurately predicting the scarce resources.

Neural network $\fnF$ transforms each scarce resource, or input vector, into a different representation for each agent. In the special case where $\fnF$ is a constant function that returns a parameter $W\suptag{mem}_{ijh}$ with learned memories, agents compete against each other to use or ignore, not representations computed from the actual scarce resources, but representations learned independently of such resources---``imagined resources,'' as it were.

From this viewpoint, ``bang per bit'' is a form of {\em utility}, and parameters $\beta\use_{ij}$ and $\beta\ign_{ij}$ are {\em net prices} each agent pays or collects per unit of data used or ignored to maximize utility, in service of a training objective specified elsewhere as a dependency of the final agent states. If no agent can improve its utility, given the state of all other agents, the competition has reached a local optimum specific to the implementation of $\fnA$, $\fnF$, $\fnG$, and $\fnS$.

\begin{figure}[t]
	\vskip 0.1in
	\begin{center}
		\centerline{\includegraphics{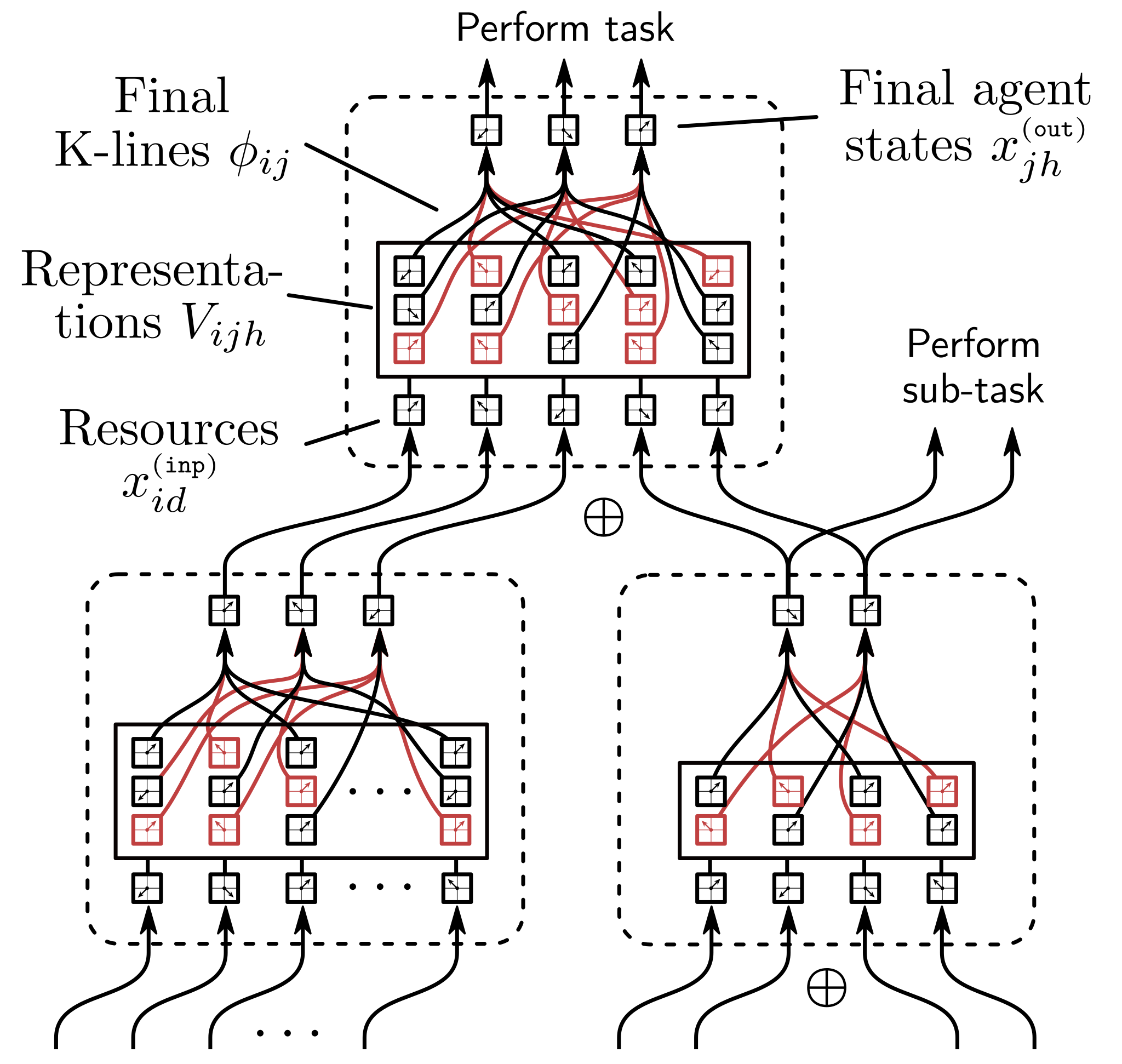}}
		\caption{A network of blocks in a model of a Society of Mind \cite{10.5555/22939}. In each block, agents use or ignore representations of resources via K-lines to perform tasks learned in training.}
		\label{fig:connected_blocks}
	\end{center}
	\vskip -0.2in
\end{figure}

The ``bang per bit'' coefficients $\phi_{ij}$ \eqref{eq:phi_ij} function as ``knowledge lines,'' or {\em K-lines}, connecting agents to representations of resources as necessary to perform tasks learned in training. If we call an instance of the algorithm a ``block,'' multiple blocks can interact with each other via their respective agents' final states, dynamically connected via K-lines to perform tasks learned in training, modeling a Society of Mind \cite{10.5555/22939}, as shown in Figure \ref{fig:connected_blocks}, but with one significant difference: In our algorithm, the agents in each block incur a net cost for ignoring their representations of the available resources.

\section{Efficient Implementation}\label{sec:implementation}

The number of possible implementations of $\fnA$, $\fnF$, $\fnG$, and $\fnS$ is infinite. Here, we present one implementation (Algorithm \ref{alg2:Efficient_Implementation}), incorporating three significant optimizations that reduce parameter count, memory use, and computation by orders of magnitude, overcoming the difficulties to routing longer sequences discussed in \ref{ssec:V_ijh}.

\begin{algorithm*}[t]
	\small
	\DontPrintSemicolon
	\caption{Our implementation of $\fnA$, $\fnF$, $\fnG$, and $\fnS$. Trivial optimizations are not shown for ease of exposition. $\mathfrak{N}$ denotes normalization of each vector's elements to zero mean and unit variance for numerical stability. If $n\inp$ is variable, we remove index $i$ from all parameters that have it and compute $\beta\use_{ij} \longleftarrow \sum_d x\inp_{id} W\use_{dj} + B\use_j$ and $\beta\ign_{ij} \longleftarrow \sum_d x\inp_{id} W\ign_{dj} + B\ign_j$.}
	\label{alg2:Efficient_Implementation}
	\KwIn{$x\inp_{id}$.}
	\KwOut{$x\out_{jh}$.}
	\BlankLine
	$a\inp_i \longleftarrow \frac{ \sum_d W\suptag{\fnA}_{id} x\inp_{id} }{ \sqrt{n\inp} } + B\suptag{\fnA}_i$ \tcp*{apply $\fnA$, a scaled linear transformation with bias per input vector $i$}
	\For{$n\iters$ iterations}{
		\Begin(E-Step){
			\eIf{on first iteration}{
				$R_{ij} \longleftarrow \frac{1}{n\out}$\;
			}{
				$\hat{x}\inp_{jd} \longleftarrow  W\suptag{\fnG_2}_{jd} \sum_h W\suptag{\fnG_1}_{hd} \mathfrak{N}(x\out_{jh}) + B\suptag{\fnG_2}_{jd}$ \tcp*{apply $\fnG$, a two-layer neural network per output vector $j$}
				$S_{ij} \longleftarrow \log f \big( W\suptag{\fnS}_{ij} \sum_d x\inp_{id} \hat{x}\inp_{jd} + B\suptag{\fnS}_{ij} \big)$ \tcp*{apply $\fnS$, a nonlinear transformation per dot-product $ij$}
				$R_{ij} \longleftarrow \frac{ e^{S_{ij}} }{ \sum_j e^{S_{ij}} }$\;
			}
		}
		\Begin(D-Step){
			$D\use_{ij} \longleftarrow f(a\inp_i) R_{ij}$\;
			$D\ign_{ij} \longleftarrow f(a\inp_i) - D\use_{ij}$\;
		}
		\Begin(M-Step){
			$\phi_{ij} \longleftarrow \beta\use_{ij} D\use_{ij} -  \beta\ign_{ij} D\ign_{ij}$ \tcp*{compute ``bang per bit'' coefficients $i$ for each basis $j$}\label{alg2:factored_out}
			$x\out_{jh} \longleftarrow \frac{ \sum_d W\suptag{\fnF_2}_{dh} W\suptag{\fnF_1}_{jd} \sum_i \phi_{ij} x\inp_{id} }{ \sqrt{n\inp} } + \sum_i \phi_{ij} B\suptag{\fnF_2}_{jh}$ \tcp*{lazily evaluate $\fnF \left(x\inp_{id}\right)$ and efficiently contract votes}\label{alg2:lazy_F}
		}
	}
\end{algorithm*}

\subsection{Efficient Implementation of $\fnF$}

Our first significant optimization is to implement $\fnF$ with orders of magnitude fewer parameters than would be necessary were we to apply a different set of linear transformations per output vector (as in the original variant of the algorithm).

We define $\fnF$ as a two-layer neural network:

\begin{equation}
\fnF(\cdot) := \fnF_2(\fnF_1(\cdot)),
\end{equation}

where

\begin{equation}
\begin{aligned}
\fnF_1(\cdot) & := \frac{ 1 }{ \sqrt{n\inp} } (\cdot) W\suptag{\fnF_1}_{jd} \\
\fnF_2(\cdot) & := \sum_d W\suptag{\fnF_2}_{dh} (\cdot) + B\suptag{\fnF_2}_{jh}.
\end{aligned}
\end{equation}

When we apply $\fnF$ to a sequence of input vectors $x\inp_{id}$, $\fnF_1$ computes a tensor product:

\begin{equation}
{\small
	\underbrace{
		\mathbb{R}^{n\inp \!\times d\inp}
	}_{x\inp_{id}}
	\!\otimes
	\underbrace{
		\mathbb{R}^{n\out \!\times d\inp}
	}_{W\suptag{\fnF_1}_{jd}}
	\!\to\!
	\underbrace{
		\mathbb{R}^{n\inp \!\times n\out \!\times d\inp}
	}_{x\inp_{id} W\suptag{\fnF_1}_{jd}},
}
\end{equation}

giving us, for each of $n\inp$ input vectors, $n\out$ different elementwise scalings, or Hadamard products, of its $d\inp$ elements. We scale the tensor product by $\frac{1}{\sqrt{n\inp}}$ to keep the subsequent contraction of votes over index $i$ in the same region for different values of $n\inp$. $\fnF_2$ applies parameter $W\suptag{\fnF_2}_{dh}$ as a linear transformation from $\mathbb{R}^{d\inp}$ to $\mathbb{R}^{d\out}$, and then adds $B\suptag{\fnF_2}_{jh}$, a different bias per output vector basis, making it possible for all bases to span up to $d\out$ dimensions when $n\inp \ge d\out$ but $d\inp < d\out$.\footnote{
	If $d\inp < d\out$ and we don't add biases to the bases, they would all span the same subspace of dimension $d\suptag{subspace} \le \min \left(n\inp, \text{rank}(W\suptag{\fnF_2}_{dh}) \right) < d\out$.
} The tensor product and per-basis biases break symmetry.

$\fnF$'s parameter count in this implementation is $n\out d\inp + d\inp d\out + n\out d\out$, versus $n\inp n\out d\inp d\out$ were we to apply $n\inp n\out$ different linear transformations, or $n\out d\inp d\out$ were we to apply $n\out$ different linear transformations to each input vector. The trade-off of this reduction in parameter count is that, for any fixed $n\inp$, $n\out$, $d\inp$, and $d\out$, the space of transformations learnable by $\fnF$ is smaller.

\subsection{Lazy Evaluation of $\fnF$}

Our second significant optimization is to evaluate $\fnF$ lazily in each iteration, in order to compute and contract votes as needed without having to store all of them simultaneously in memory as intermediate values: The tensor $V_{ijh}$ disappears from all expressions. Only the output vectors need be stored at the end of each iteration (Algorithm \ref{alg2:Efficient_Implementation}, line \ref{alg2:lazy_F}). The lazy evaluation of $\fnF$ and immediate contraction of each vote can be done efficiently, {\em i.e.}, in parallel, because our implementation of $\fnF$ computes each input vector's vote for each output vector independently from every other vote.

By never storing votes in a tensor $V_{ijh}$, we reduce memory footprint by $\bigO(n\inp n\out d\out)$. The trade-off of this reduction in footprint is an increase in computation that is linear in the number of iterations: We now compute all votes in every iteration, instead of only once before the loop.

\subsection{Efficient Evaluation of $\fnF$}

Our third significant optimization is necessary to avoid having to store intermediate-value tensors with $n\inp \times n\out \times d\inp$ or $n\inp \times n\out \times d\out$ elements simultaneously in memory, and also to avoid computing votes twice in each iteration, which is a side effect of the lazy evaluation of $\fnF$, due to our computation of output vectors as a difference of two weighted sums of votes (Algorithm \ref{alg1:General_Formulation}, line \ref{alg1:x_out}), both now lazily evaluated.

We factorize the difference of weighted sums into the tensor contraction $\sum_i \phi_{ij} \fnF_2(\fnF_1(x\inp_{id}))$, where $\phi_{ij}$ are the ``bang per bit'' coefficients (Algorithm \ref{alg2:Efficient_Implementation}, line \ref{alg2:factored_out}), and algebraically manipulate it to obtain the expression in Algorithm \ref{alg2:Efficient_Implementation}, line \ref{alg2:lazy_F}. The expression computes, weights, and contracts votes in a memory-efficient manner in each iteration, and then applies $\fnF_2$ as a last step, {\em after} contracting all votes, reducing the number of linear transformations executed in parallel by a factor of $n\inp$. See appendix \ref{app:efficient_lazy_contraction_votes} for the derivation.

\begin{figure*}[t]
	\vskip 0.1in
	\begin{center}
		\centerline{\includegraphics{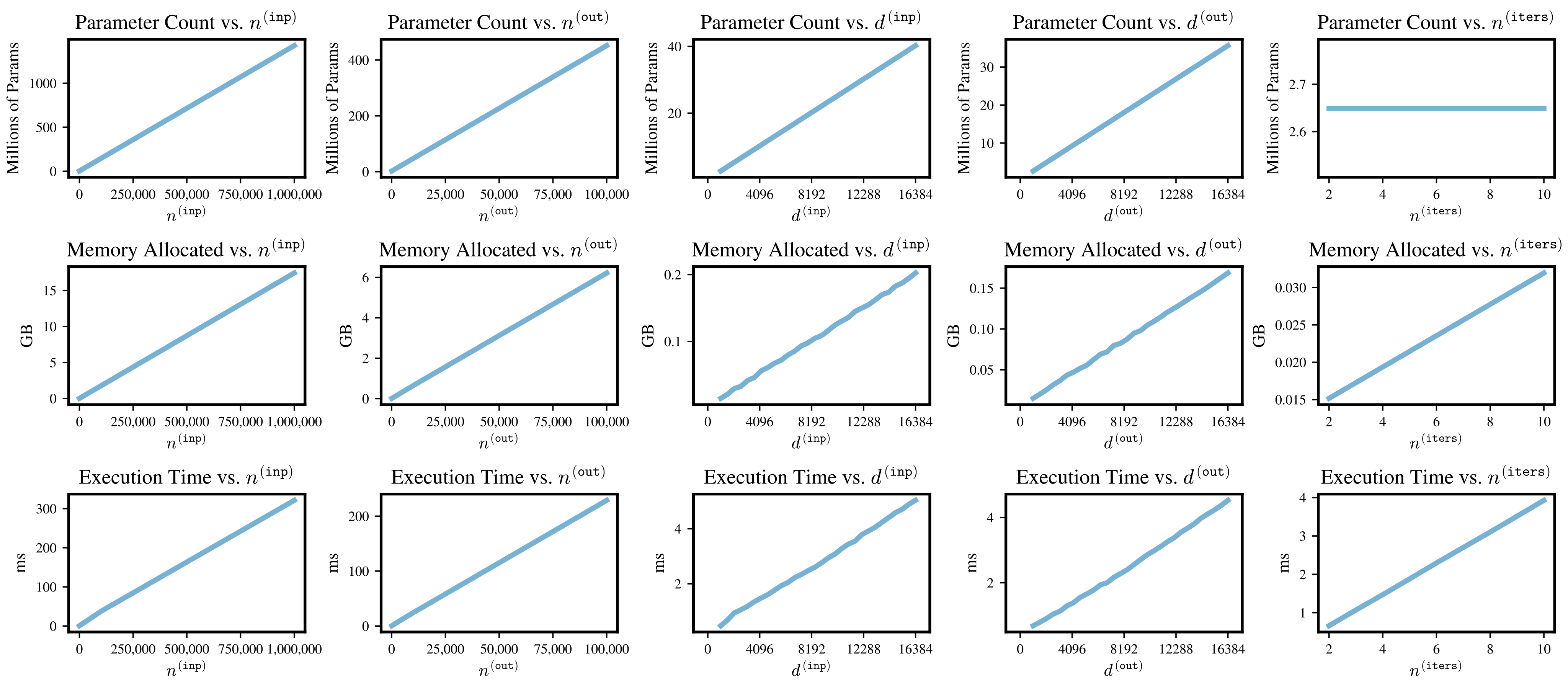}}
		\caption{Parameter count, memory footprint, and execution time of a forward pass as we vary each of $n\inp$, $n\out$, $d\inp$, $d\out$, and $n\iters$, while keeping the others constant at a baseline, at 32-bit precision, keeping track of gradients, on a recent hardware accelerator (GPU). Baseline values are 100, 100, 1024, 1024, and 2, respectively. Memory figures are peak allocations.}
		\label{fig:parameters_memory_execution}
	\end{center}
	\vskip -0.2in
\end{figure*}

\section{Experiments}\label{sec:experiments}

\subsection{Efficiency and Scalability}

We measure our implementation's parameter count, memory footprint, and execution time as we increase $n\inp$ from 100 to 1,000,000 input vectors, $n\out$ from 100 to 100,000 output vectors, $d\inp$ from 1024 to 16384 elements per input vector, $d\out$ from 1024 to 16384 elements per output vector, and number of iterations $n\iters$ from 2 to 10. We find that parameter count, memory footprint, and execution time are linear in each of $n\inp$, $n\out$, $d\inp$, and $d\out$ (Figure \ref{fig:parameters_memory_execution}), enabling fine-grained control over memory consumption and computational cost. Given a memory and compute budget, we can increase the maximum length of input sequences our implementation can route by reducing output sequence length, and vice versa. Memory footprint and execution time are also linear in the number of iterations.

\begin{figure*}[t]
	\vskip 0.1in
	\begin{center}
		\centerline{\includegraphics{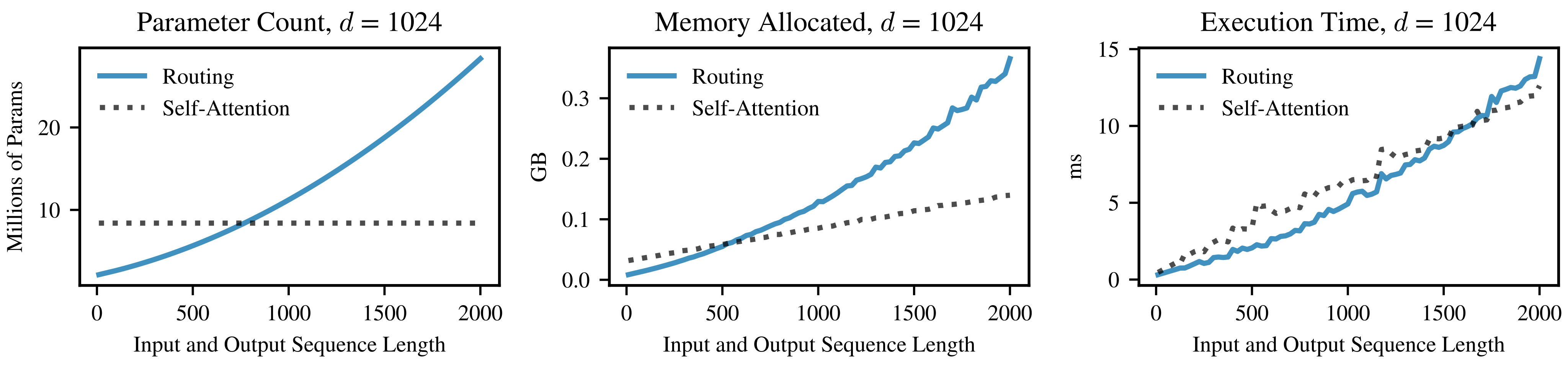}}
		\caption{Comparison to a Transformer encoder layer using self-attention. To make comparison possible, we restrict our implementation to $n\inp = n\out$, $d\inp = d\out$, and $n\iters = 2$. Data is for a forward pass on a recent hardware accelerator (GPU) at 32-bit precision, keeping track of gradients, using dense Softmax functions. Self-attention uses eight heads, the default. Memory figures are peak allocations.}
		\label{fig:comparison_to_self_attention}
	\end{center}
	\vskip -0.2in
\end{figure*}

We also compare our implementation's parameter count, memory footprint, and execution time to those of a Transformer encoder layer using self-attention as we increase sequence length up to 2000 vectors, keeping vector size constant at 1024. To make the comparison possible, we restrict our implementation to input and output sequences that have the same shape, routing over two iterations, the fewest possible. We find our implementation requires fewer parameters for sequences with up to 600 vectors, allocates less memory for sequences with up to 800 vectors, and incurs less computation for sequences with up to 1700 vectors (Figure \ref{fig:comparison_to_self_attention}), which is surprising to us, because our algorithm proposes $n\out$ output vectors per input vector, whereas the query-key-value mechanism of self-attention proposes only one output vector (a ``value'') per input vector.

\subsection{Performance on Benchmarks}

We test our implementation on six classification benchmarks in natural language and vision, obtaining accuracy that is competitive with, and in one case better than, the state of the art (Table \ref{tab:accuracies}). For each benchmark, we add a classification head to a pretrained Transformer. The head accepts as input all token embeddings computed by every Transformer layer, flattens them into a single sequence, and sequentially applies three routings:

\begin{center}
	\vskip 1em
	\begin{tabular}{l|cccc}
		               & $n\inp$ & $n\out$         & $d\inp$ & $d\out$ \\
		\midrule
		$\routing_1$   & --      & $n\hid$         & $d\emb$ & $d\hid$ \\
		$\routing_2$   & $n\hid$ & $n\hid$         & $d\hid$ & $d\hid$ \\
		$\routing_3$   & $n\hid$ & $n\suptag{cls}$ & $d\hid$ & 1       \\
	\end{tabular} \\
	\vskip 1em
\end{center}

where $\routing_1$'s number of input vectors is unspecified because the flattened sequence's length is variable, $n\hid$ is a number of hidden explanatory vectors of our choosing, $d\emb$ is the pretrained Transformer's embedding size, $d\hid$ is the size of the hidden explanatory vectors, and $n\suptag{cls}$ is the number of classes specific to each task.

For natural language tasks, we use RoBERTa-large \cite{DBLP:journals/corr/abs-1907-11692} as the pretrained Transformer. For visual tasks, we use BEiT-large with 16$\times$16 patches from 224$\times$224 images \cite{hangbo2021beit}. We freeze the Transformer. For all tasks, we specify $n\hid = 64$ and $d\hid = d\emb$. All routings execute $n\iters = 2$ iterations, the fewest possible. Before flattenning the hidden embeddings we apply layer normalization at each level of depth. If the input sequence's length is greater than the Transformer's maximum sequence length, we split the input sequence into chunks, apply the Transformer to each chunk, and join the hidden states computed for all chunks at every level of Transformer depth. The longest flattened sequence we see among all benchmarks has 89,600 input vectors, computed by RoBERTa-large's 25 hidden layers for a natural language sample drawn from the IMDB movie review dataset, split in 7 chunks, each with 512 subword tokens.

\begin{table}[h]
	\begin{center}
		\small
		\vskip 0.1in
		\begin{tabular}{llc}
			\toprule
			Classification Benchmark        &      & Accuracy (\%) \\
			\midrule
			\em Natural Language            &      &        \\
			IMDB                            &      &   96.2 \\
			SST-5*                          &      &   59.8 \\
			SST-2                           &      &   96.0 \\
			\midrule
			\em Vision                      &      &        \\
			ImageNet-1K @ 224$\times$224    & Top1 &   86.7 \\
			                                & Top5 &   98.1 \\
			CIFAR-100                       &      &   93.8 \\
			CIFAR-10                        &      &   99.2 \\
			\bottomrule
			\multicolumn{2}{l}{\small * New state-of-the-art accuracy.}
		\end{tabular}
        \caption{\label{tab:accuracies}Classification accuracy.}
		\vskip 0.25in
	\end{center}
\end{table}

\subsection{End-to-End Credit Assignments}

Vectors remain independent of each other between each routing executed in the classification head, so we can compute end-to-end credit assignments for all benchmark tasks. Each head executes three routings, giving us three credit assignment matrices. We multiply them as described in Appendix \ref{app:composability_of_credit_assignments}, obtaining a matrix of end-to-end credit assigned to every hidden Transformer embedding $i$ for each predicted classification score $j$:

\begin{equation}
\phi\suptag{e2e}_{ij}  \longleftarrow \frac{
	\sum_{j'j''} \phi\suptag{\routing_1}_{ij'} \phi\suptag{\routing_2}_{j'j''} \phi\suptag{\routing_3}_{j''j}
}{
	\sigma \left( \sum_{j'j''} \phi\suptag{\routing_1}_{ij'} \phi\suptag{\routing_2}_{j'j''} \phi\suptag{\routing_3}_{j''j} \right)
},
\end{equation}

where $j' = (1, 2, \dots, n\hid)$ and $j'' = (1, 2, \dots, n\hid)$, and $\sigma$ computes the standard deviation over all elements, scaling the credit assignments to unit variance. The largest end-to-end credit assignment matrix we see among all benchmarks has 4925$\times$1000 elements, consisting of the end-to-end credit assigned to embeddings of a special token and 196 image patches computed by each of BEiT-large's 25 levels of depth, in a flattened sequence with 4925 input vectors, for 1000 predicted scores, each an output vector with one element, for ImageNet-1K classification.

\begin{figure*}[t]
	\vskip 0.1in
	\begin{center}
		\centerline{\includegraphics{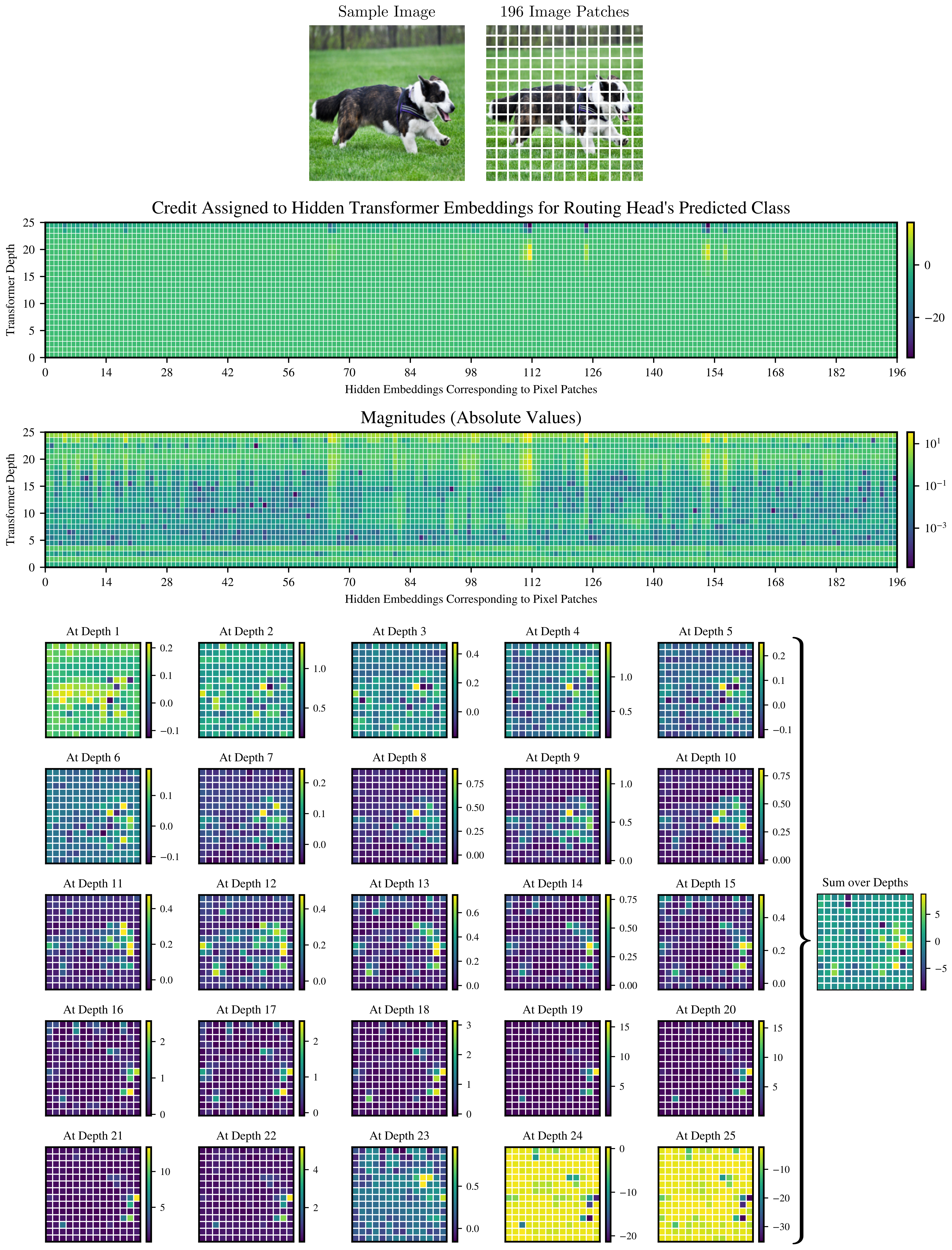}}
		\caption{Typical example of end-to-end credit assigned to Transformer hidden states in a visual task. Here, our three-layer routing head assigns credit to the dog's entire body in shallower layers, and to its nose, mouth, ears, and paws in deeper layers. The matrix of end-to-end credit assignments $\phi\suptag{e2e}_{ij}$ has 4925$\times$1000 elements, consisting of credit assigned to 197 hidden embeddings at 25 levels of Transformer depth, or 4925 input vectors, for 1000 classification scores, each an output vector with one element. We show the absolute values of 4900 credit assignments to embeddings corresponding to 196 image patches, for the highest score, excluding 25 credit assignments to a special token added to the input sequence.}
		\label{fig:sample_credit_assignments_vision}
	\end{center}
	\vskip -0.2in
\end{figure*}

\begin{figure*}[t]
	\vskip 0.1in
	\begin{center}
		\centerline{\includegraphics{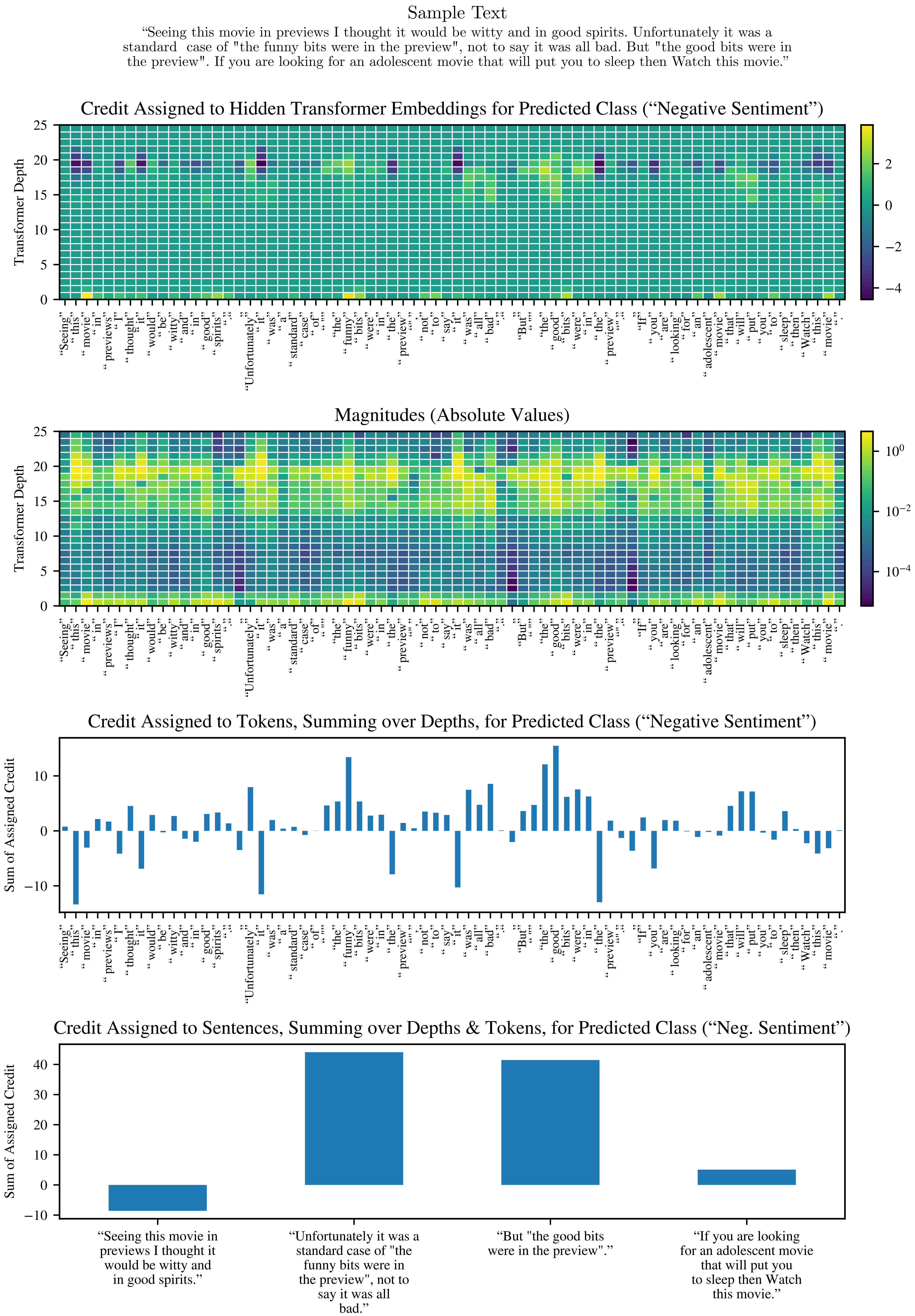}}
		\caption{Typical example of end-to-end credit assigned to Transformer hidden states in a natural language task. Here, $\phi\suptag{e2e}_{ij}$ has 1850$\times$2 elements, consisting of credit assigned to 74 hidden embeddings at 25 levels of Transformer depth, or 1850 input vectors, for 2 classification scores, each an output vector with one element. We show 1800 credit assignments to embeddings corresponding to 72 subword tokens for the highest score, excluding 50 credit assignments to two special tokens added to the input sequence.}
		\label{fig:sample_credit_assignments_natural_language}
	\end{center}
	\vskip -0.2in
\end{figure*}

We sum $\phi\suptag{e2e}_{ij}$'s elements over all levels of Transformer depth to obtain the credit assigned to subword tokens and pixel patches, and over groups of tokens and patches to obtain the credit assigned to sentences and image regions. We find the end-to-end credit assignments are interpretable. Figures \ref{fig:sample_credit_assignments_vision} and \ref{fig:sample_credit_assignments_natural_language} show typical examples.

\cleardoublepage

\bibliography{An_Algorithm_for_Routing_Vectors_in_Sequences}

\begin{thebibliography}{}
\expandafter\ifx\csname natexlab\endcsname\relax\def\natexlab#1{#1}\fi

\bibitem[{Bai et~al.(2019)Bai, Kolter, and
  Koltun}]{DBLP:journals/corr/abs-1909-01377}
Shaojie Bai, J.~Zico Kolter, and Vladlen Koltun. 2019.
\newblock Deep equilibrium models.
\newblock {\em CoRR\/} abs/1909.01377.

\bibitem[{Bao et~al.(2021)Bao, Dong, and Wei}]{hangbo2021beit}
Hangbo Bao, Li~Dong, and Furu Wei. 2021.
\newblock Beit: {BERT} pre-training of image transformers.
\newblock {\em CoRR\/} abs/2106.08254.

\bibitem[{Dou et~al.(2019)Dou, Tu, Wang, Wang, Shi, and
  Zhang}]{DBLP:journals/corr/abs-1902-05770}
Zi{-}Yi Dou, Zhaopeng Tu, Xing Wang, Longyue Wang, Shuming Shi, and Tong Zhang.
  2019.
\newblock Dynamic layer aggregation for neural machine translation with
  routing-by-agreement.
\newblock {\em CoRR\/} abs/1902.05770.

\bibitem[{Hahn et~al.(2019)Hahn, Pyeon, and Kim}]{NEURIPS2019_e46bc064}
Taeyoung Hahn, Myeongjang Pyeon, and Gunhee Kim. 2019.
\newblock Self-routing capsule networks.
\newblock In H.~Wallach, H.~Larochelle, A.~Beygelzimer, F.~d\textquotesingle
  Alch\'{e}-Buc, E.~Fox, and R.~Garnett, editors, {\em Advances in Neural
  Information Processing Systems\/}. Curran Associates, Inc., volume~32.

\bibitem[{Heinsen(2019)}]{DBLP:journals/corr/abs-1911-00792}
Franz~A. Heinsen. 2019.
\newblock An algorithm for routing capsules in all domains.
\newblock {\em CoRR\/} abs/1911.00792.

\bibitem[{Hinton et~al.(2018)Hinton, Sabour, and Frosst}]{46653}
Geoffrey Hinton, Sara Sabour, and Nicholas Frosst. 2018.
\newblock Matrix capsules with em routing.
\newblock In {\em International Conference on Learning Representations
  (ICLR)\/}.

\bibitem[{Krotov and Hopfield(2021)}]{krotov2021large}
Dmitry Krotov and John Hopfield. 2021.
\newblock Large associative memory problem in neurobiology and machine
  learning.
\newblock {\em CoRR\/} abs/1710.09829.

\bibitem[{Liu et~al.(2019)Liu, Ott, Goyal, Du, Joshi, Chen, Levy, Lewis,
  Zettlemoyer, and Stoyanov}]{DBLP:journals/corr/abs-1907-11692}
Yinhan Liu, Myle Ott, Naman Goyal, Jingfei Du, Mandar Joshi, Danqi Chen, Omer
  Levy, Mike Lewis, Luke Zettlemoyer, and Veselin Stoyanov. 2019.
\newblock Roberta: {A} robustly optimized {BERT} pretraining approach.
\newblock {\em CoRR\/} abs/1907.11692.

\bibitem[{Lundberg and Lee(2017)}]{NIPS2017_7062}
Scott~M Lundberg and Su-In Lee. 2017.
\newblock A unified approach to interpreting model predictions.
\newblock In I.~Guyon, U.~V. Luxburg, S.~Bengio, H.~Wallach, R.~Fergus,
  S.~Vishwanathan, and R.~Garnett, editors, {\em Advances in Neural Information
  Processing Systems 30\/}, Curran Associates, Inc., pages 4765--4774.

\bibitem[{Minsky(1986)}]{10.5555/22939}
Marvin Minsky. 1986.
\newblock {\em The Society of Mind\/}.
\newblock Simon and Schuster, Inc., USA.

\bibitem[{Rajasegaran et~al.(2019)Rajasegaran, Jayasundara, Jayasekara,
  Jayasekara, Seneviratne, and Rodrigo}]{DBLP:journals/corr/abs-1904-09546}
Jathushan Rajasegaran, Vinoj Jayasundara, Sandaru Jayasekara, Hirunima
  Jayasekara, Suranga Seneviratne, and Ranga Rodrigo. 2019.
\newblock Deepcaps: Going deeper with capsule networks.
\newblock {\em CoRR\/} abs/1904.09546.

\bibitem[{Ramsauer et~al.(2021)Ramsauer, Schäfl, Lehner, Seidl, Widrich,
  Adler, Gruber, Holzleitner, Pavlović, Sandve, Greiff, Kreil, Kopp,
  Klambauer, Brandstetter, and Hochreiter}]{ramsauer2021hopfield}
Hubert Ramsauer, Bernhard Schäfl, Johannes Lehner, Philipp Seidl, Michael
  Widrich, Thomas Adler, Lukas Gruber, Markus Holzleitner, Milena Pavlović,
  Geir~Kjetil Sandve, Victor Greiff, David Kreil, Michael Kopp, Günter
  Klambauer, Johannes Brandstetter, and Sepp Hochreiter. 2021.
\newblock Hopfield networks is all you need.
\newblock {\em CoRR\/} abs/2008.02217.

\bibitem[{Ribeiro et~al.(2020)Ribeiro, Leontidis, and
  Kollias}]{ribeiro2020capsule}
Fabio De~Sousa Ribeiro, Georgios Leontidis, and Stefanos~D Kollias. 2020.
\newblock Capsule routing via variational bayes.
\newblock In {\em AAAI\/}. pages 3749--3756.

\bibitem[{Sabour et~al.(2017)Sabour, Frosst, and
  Hinton}]{DBLP:journals/corr/abs-1710-09829}
Sara Sabour, Nicholas Frosst, and Geoffrey~E. Hinton. 2017.
\newblock Dynamic routing between capsules.
\newblock {\em CoRR\/} abs/1710.09829.

\bibitem[{Tsai et~al.(2020)Tsai, Srivastava, Goh, and
  Salakhutdinov}]{tsai2020Capsules}
Yao-Hung~Hubert Tsai, Nitish Srivastava, Hanlin Goh, and Ruslan Salakhutdinov.
  2020.
\newblock Capsules with inverted dot-product attention routing.
\newblock In {\em International Conference on Learning Representations
  (ICLR)\/}.

\bibitem[{Vaswani et~al.(2017)Vaswani, Shazeer, Parmar, Uszkoreit, Jones,
  Gomez, Kaiser, and Polosukhin}]{DBLP:journals/corr/VaswaniSPUJGKP17}
Ashish Vaswani, Noam Shazeer, Niki Parmar, Jakob Uszkoreit, Llion Jones,
  Aidan~N. Gomez, Lukasz Kaiser, and Illia Polosukhin. 2017.
\newblock Attention is all you need.
\newblock {\em CoRR\/} abs/1706.03762.

\bibitem[{Wang and Liu(2018)}]{wang2018an}
Dilin Wang and Qiang Liu. 2018.
\newblock An optimization view on dynamic routing between capsules.
\newblock In {\em International Conference on Learning Representations
  (ICLR)\/}.

\bibitem[{Xinyi and Chen(2019)}]{xinyi2018capsule}
Zhang Xinyi and Lihui Chen. 2019.
\newblock Capsule graph neural network.
\newblock In {\em International Conference on Learning Representations
  (ICLR)\/}.

\bibitem[{Zhang et~al.(2018)Zhang, Zhao, Wu, and
  Zhou}]{DBLP:journals/corr/abs-1805-10807}
Suofei Zhang, Wei Zhao, Xiaofu Wu, and Quan Zhou. 2018.
\newblock Fast dynamic routing based on weighted kernel density estimation.
\newblock {\em CoRR\/} abs/1805.10807.

\end{thebibliography}
\bibliographystyle{An_Algorithm_for_Routing_Vectors_in_Sequences}

\cleardoublepage

\appendix

\section{Composability of Credit Assignments}\label{app:composability_of_credit_assignments}

If each input vector's votes are independent of other input vectors' votes, then we can compose the ``bang per bit'' credit-assignment coefficients $\phi_{ij}$ on their own, independently of data transformations. Here, we show methods for computing end-to-end credit assignments over four common compositions of routings.\footnote{Subject to the same condition of independence, our methods apply also to modern Hopfield networks with bipartite structure, including Transformer self-attention, as they are simplifications of our routing algorithm. See \ref{ssec:as_associative_memory_states}.}

\subsection{In Sequential Routings}

If we compose the sequential application of two routings, $\routing_1$ and $\routing_2$, into a neural network,

\begin{equation}
x\out_{jh} \longleftarrow \routing_2(\routing_1(x\inp_{id})),
\end{equation}

we can obtain the neural network's end-to-end credit assignments $\phi\suptag{e2e}_{ij}$, over both routings, by multiplying their credit-assignment matrices,

\begin{equation}
\phi\suptag{e2e}_{ij} \longleftarrow \sum_{j'} \phi\suptag{\routing_1}_{ij'} \phi\suptag{\routing_2}_{j'j},
\end{equation}

where $\phi\suptag{\routing_1}_{ij'}$ and $\phi\suptag{\routing_2}_{j'j}$ are the credit-assignment matrices computed by $\routing_1$ and $\routing_2$, respectively, and $j'$ is the common index over $\routing_1$'s output vectors and $\routing_2$'s input vectors.

For a longer sequence of routings, we can obtain end-to-end credit assignments by multiplying the corresponding chain of credit-assignment matrices, as matrix multiplication is associative.

\subsection{In Residual Routings}

If we apply one routing as a residual to another,

\begin{equation}
x\out_{jh} \longleftarrow \routing_1(x\inp_{id}) + \routing_2(\routing_1(x\inp_{id})),
\end{equation}

we can obtain end-to-end credit assignments $\phi\suptag{e2e}_{ij}$ by adding the product of the two credit-assignment matrices to the first one,

\begin{equation}
\phi\suptag{e2e}_{ij} \longleftarrow \phi\suptag{\routing_1}_{ij} + \sum_{j'} \phi\suptag{\routing_1}_{ij'} \phi\suptag{\routing_2}_{j'j},
\end{equation}

where $\phi\suptag{\routing_1}_{ij}$ and $\phi\suptag{\routing_2}_{j'j}$ are the credit assignment matrices computed by $\routing_1$ and $\routing_2$, respectively, and $j'=j$ (necessary for disambiguation).

For a sequence of residual routings, we can obtain end-to-end credit assignments by multiplying each additional residual credit-assignment matrix with, and then adding the result back to, the previous state of the end-to-end credit-assignment matrix, as matrix addition is associative.

\subsection{In Sums of Routings}

If we sum two independent routings $\routing_1$ and $\routing_2$,

\begin{equation}
x\out_{jh} \longleftarrow \routing_1(x\suptag{inp1}_{i_1 d_1}) + \routing_2(x\suptag{inp2}_{i_2 d_2}),
\end{equation}

we can obtain end-to-end credit assignments by concatenating the two credit-assignment matrices over their mutually exclusive indices,

\begin{equation}
\phi\suptag{e2e}_{ij} \longleftarrow \phi\suptag{\routing_1}_{i_1 j} \oplus \phi\suptag{\routing_2}_{i_2 j} = 
\begin{bmatrix}
	\phi\suptag{\routing_1}_{i_1 j} \\[0.5em]
	\phi\suptag{\routing_2}_{i_2 j} \\
\end{bmatrix},
\end{equation}

where $\phi\suptag{\routing_1}_{i_1 j}$ and $\phi\suptag{\routing_2}_{i_2 j}$ are the credit assignment matrices computed by $\routing_1$ and $\routing_2$, respectively, the symbol $\oplus$ denotes a direct sum over mutually exclusive indices $i_1$ and $i_2$, and $i = (i_1; i_2)$ is a single index that concatenates indices $i_1$ and $i_2$.

For sums of three or more independent routings, we can obtain end-to-end credit assignments by concatenating their credit-assignment matrices over the mutually exclusive input indices, but we must fix the order of concatenation, as direct sums are associative but not commutative.

\subsection{In Concatenations of Routings}

If we concatenate the output vectors of two independent routings $\routing_1$ and $\routing_2$,

\begin{equation}
\begin{aligned}
x\suptag{hid1}_{j_1 h} \longleftarrow & \; \routing_1(x\suptag{inp1}_{i_1 d_1}) \\
x\suptag{hid2}_{j_2 h} \longleftarrow & \; \routing_2(x\suptag{inp2}_{i_2 d_2}) \\
x\out_{jh} \longleftarrow & \; x\suptag{hid1}_{j_1h} \oplus x\suptag{hid2}_{j_2 h},\\
\end{aligned}
\end{equation}

where $j = (j_1; j_2)$ is the concatenated index, we can obtain end-to-end credit assignments with a direct sum of the credit-assignment matrices,

\begin{equation}
\phi\suptag{e2e}_{ij} \longleftarrow \phi\suptag{\routing_1}_{i_1 j_1} \oplus \phi\suptag{\routing_2}_{i_2 j_2}
= \begin{bmatrix}
	\phi\suptag{\routing_1}_{i_1 j_1} & 0                                 \\[0.5em]
	0                                 & \phi\suptag{\routing_2}_{i_2 j_2} \\
\end{bmatrix},
\end{equation}

where $\phi\suptag{\routing_1}_{i_1 j_1}$ and $\phi\suptag{\routing_2}_{i_2 j_2}$ are the credit assignment matrices computed by $\routing_1$ and $\routing_2$, respectively, $\oplus$ again denotes a direct sum over mutually exclusive indices, $i = (i_1; i_2)$, and $j = (j_1; j_2)$.

For three or more concatenations, we can obtain end-to-end credit assignments with direct sums over mutually exclusive indices, provided we fix the order of concatenation, as direct sums are associative but not commutative.

\cleardoublepage

\section{Derivation of Update Rule}\label{app:derivation_of_update_rule}

\vspace{-0.3in}
\begin{multicols}{1}
	\begin{equation*}
	\begin{aligned}
	\fnU( \; \cdot \; |x\inp_{id}) 
	= & \sum_i \beta\use_{ij} D\use_{ij} V_{ijh} - \sum_i \beta\ign_{ij} D\ign_{ij} V_{ijh}
	\\
	= & \sum_i \left( \beta\use_{ij} D\use_{ij} V_{ijh} - \beta\ign_{ij} D\ign_{ij} V_{ijh} \right)
	\\
	= & \sum_i \left( \beta\use_{ij} \left( f(a\inp_i) R_{ij} \right) V_{ijh} - \beta\ign_{ij} \left( f(a\inp_i) - f(a\inp_i) R_{ij} \right) V_{ijh} \right)
	\\
	= & \sum_i \left( \beta\use_{ij} f(a\inp_i) R_{ij} V_{ijh} - \beta\ign_{ij} f(a\inp_i) V_{ijh} + \beta\ign_{ij} f(a\inp_i) R_{ij} V_{ijh} \right)
	\\
	= & \sum_i \left( R_{ij} ( \beta\use_{ij} + \beta\ign_{ij} ) f(a\inp_i) V_{ijh} - \beta\ign_{ij} f(a\inp_i) V_{ijh} \right)
	\\
	= & \sum_i \Big( R_{ij} \underbrace{ ( \beta\use_{ij} \! + \! \beta\ign_{ij} ) f(\fnA(x\inp_{id})) \fnF(x\inp_{id}) }_{ \textcomment{Define as } \fnM(x\inp_{id}) }
	- \underbrace{ \beta\ign_{ij} f(\fnA(x\inp_{id})) \fnF(x\inp_{id}) }_{ \textcomment{Define as } \fnB(x\inp_{id}) } \Big)
	\\
	= & \sum_i \! \Bigg( \! \underbrace{ \frac{ e^{\fnS(x\inp_{id},\, \fnG( \cdot ))} }{ \sum_j \! e^{\fnS(x\inp_{id},\, \fnG( \cdot ))} } }_{ \textcomment{Define as } \fnR( \; \cdot \; |x\inp_{id}) }  \fnM(x\inp_{id}) - \fnB(x\inp_{id}) \! \Bigg)
	\\
	= & \sum_i \Big( \fnR ( \; \cdot \; | \underbrace{ x\inp_{id} }_{\textcomment{Keys}} ) \underbrace{ \fnM(x\inp_{id}) }_{\textcomment{Values}} - \underbrace{ \fnB(x\inp_{id}) }_{\textcomment{Biases}} \Big)
	\quad \begin{subarray}{l}
	\textcomment{// $\fnR$ computes each iteration's {\em routing probabilities}.} \\
	\textcomment{// $\fnM$ obtains content-addressable {\em memory values}.} \\
	\textcomment{// $\fnB$ obtains content-addressable {\em memory biases}.} \\
	\end{subarray}
	\\
	\end{aligned}
	\end{equation*}
\end{multicols}
\vskip 0.05in

\section{Efficient Lazy Contraction of Votes}\label{app:efficient_lazy_contraction_votes}

\vspace{-0.3in}
\begin{multicols}{1}
	\begin{equation*}
	\begin{aligned}
	\sum_i \phi_{ij} V_{ijh}
	& = \sum_i \phi_{ij} \fnF_2 \left(\fnF_1 \left( x\inp_{id} \right) \right)
	& & \textcomment{// Lazy evaluation in each iteration.}
	\\
	& = \sum_i \phi_{ij}  \Bigg( \underbrace{ \underbrace{ \sum_d W\suptag{\fnF_2}_{dh} \underbrace{ \left( \frac{ \quad x\inp_{id} W\suptag{\fnF_1}_{jd} \quad }{ \sqrt{n\inp} } \right) }_{ \bigO(n\inp \times n\out \times d\inp) } }_{ \bigO(n\inp \times n\out \times d\out) }  + B\suptag{\fnF_2}_{jh} }_{ \bigO(n\inp \times n\out \times d\out) } \Bigg)
	& & \begin{subarray}{l}
	\textcomment{// If we evaluate expression naively,} \\
	\textcomment{// all intermediate tensors occupy} \\
	\textcomment{// either $\bigO(n\inp n\out d\inp)$ or} \\
	\textcomment{// $\bigO(n\inp n\out d\out)$ space.} \\
	\end{subarray}
	\\
	& = \sum_i \phi_{ij} \sum_d W\suptag{\fnF_2}_{dh} \frac{ x\inp_{id} W\suptag{\fnF_1}_{jd} }{ \sqrt{n\inp} } + \sum_i \phi_{ij} B\suptag{\fnF_2}_{jh}
	& & \textcomment{// Distribute contraction with $\phi_{ij}$.}
	\\
	& = \frac{ 1 }{ \sqrt{n\inp} } \sum_i \phi_{ij} \sum_d W\suptag{\fnF_2}_{dh} x\inp_{id} W\suptag{\fnF_1}_{jd} + \sum_i \phi_{ij} B\suptag{\fnF_2}_{jh}
	& & \textcomment{// Factor out scalar in first term.}
	\\
	& = \frac{ 1 }{ \sqrt{n\inp} } \sum_{id} W\suptag{\fnF_2}_{dh} W\suptag{\fnF_1}_{jd} \phi_{ij} x\inp_{id} + \sum_i \phi_{ij} B\suptag{\fnF_2}_{jh}
	& & \begin{subarray}{l}
	\textcomment{// Express first term as a sequence} \\
	\textcomment{// of elementwise tensor operations} \\
	\textcomment{// contracted over two indices, $id$.} \\
	\end{subarray}
	\\
	& = \frac{ 1 }{ \sqrt{n\inp} } \sum_d W\suptag{\fnF_2}_{dh} W\suptag{\fnF_1}_{jd} \sum_i \phi_{ij} x\inp_{id} + \sum_i \phi_{ij} B\suptag{\fnF_2}_{jh}
	& & \begin{subarray}{l}
	\textcomment{// Contract over $i$ before multiplying} \\
	\textcomment{// elementwise by $W\suptag{\fnF_1}_{jd}$.} \\
	\end{subarray}
	\\
	& = \frac{ 1 }{ \sqrt{n\inp} } \underbrace{ \sum_d W\suptag{\fnF_2}_{dh} \underbrace{ \Bigg( W\suptag{\fnF_1}_{jd} \underbrace{ \Bigg( \sum_i \phi_{ij} x\inp_{id} \Bigg) }_{ \bigO(n\out \times d\inp) } \Bigg) }_{ \bigO(n\out \times d\inp) } }_{ \bigO(n\out \times d\out) } + \underbrace{ \sum_i \phi_{ij} B\suptag{\fnF_2}_{jh} }_{ \bigO(n\out \times d\out) }
	& & \begin{subarray}{l}
	\textcomment{// Now, all intermediate tensors} \\
	\textcomment{// occupy either $\bigO(n\out d\inp)$} \\
	\textcomment{// or $\bigO(n\out d\out)$ space, and} \\
	\textcomment{// we apply $\fnF_2$ as a last step} \\
	\textcomment{// only once per output vector.} \\
	\end{subarray}
	\\
	\end{aligned}
	\end{equation*}
\end{multicols}
\vskip 0.05in

\end{document}